\documentclass[10pt,journal,compsoc]{IEEEtran}
\ifCLASSINFOpdf
\else
\fi
\usepackage{amsmath,amsfonts}
\usepackage{algorithmic}
\usepackage{algorithm}
\usepackage{array}
\usepackage{subcaption}
\usepackage{textcomp}
\usepackage{stfloats}
\usepackage{url}
\usepackage{verbatim}
\usepackage{graphicx}
\usepackage{cite}
\usepackage{makecell}
\usepackage{multirow}
\usepackage[monochrome]{xcolor}
\usepackage{enumitem}

\usepackage[]{mdframed}

\usepackage[bookmarks=true, hidelinks]{hyperref}

\newtheorem{definition}{Definition}

\def\x{\mathbf{x}}
\def\y{\mathbf{y}}
\def\z{\mathbf{z}}
\def\w{\mathbf{w}}

\def\ie{{\em i.e.}}
\def\eg{{\em e.g.}}
\let\oldemptyset\emptyset
\let\emptyset\varnothing

\newcommand{\shu}{\textcolor{red}}

\begin{document}
%
\title{Rank-based Decomposable Losses in Machine Learning: A Survey}
%
%
%
%

\author{Shu Hu$^{*}$, Xin Wang,~\IEEEmembership{Senior Member,~IEEE,} 
        Siwei Lyu,~\IEEEmembership{Fellow,~IEEE}

\IEEEcompsocitemizethanks{
\IEEEcompsocthanksitem Shu Hu is with the Department of Computer Information and Graphics Technology, Purdue School of Engineering and Technology at Indiana University-Purdue University Indianapolis, IN, 46202, USA and
the Heinz College of Information
Systems and Public Policy, Carnegie Mellon University, Pittsburgh, PA, 15213, USA. e-mail:(shuhu@cmu.edu)
\IEEEcompsocthanksitem Xin Wang, and Siwei Lyu are with the Department of Computer Science and
Engineering, University at Buffalo, SUNY, Buffalo, NY 14260, USA. e-mail:(\{xwang264, siweilyu\}@buffalo.edu)
\IEEEcompsocthanksitem This work was supported by NSF IIS-2008532.
}
\thanks{$^*$ Shu Hu is the corresponding author.}
}

%
%

\markboth{Journal of \LaTeX\ Class Files,~Vol.~14, No.~8, August~2022}%
{Shell \MakeLowercase{\textit{et al.}}: Bare Demo of IEEEtran.cls for Computer Society Journals}
%



\IEEEtitleabstractindextext{%
\begin{abstract}
Recent works have revealed an essential paradigm in designing loss functions that differentiate individual losses vs. aggregate losses. The individual loss measures the quality of the model on a sample, while the aggregate loss combines individual losses/scores over each training sample. Both have a common procedure that aggregates a set of individual values to a single numerical value. The ranking order reflects the most fundamental relation among individual values in designing losses. In addition, decomposability, in which a loss can be decomposed into an ensemble of individual terms, becomes a significant property of organizing losses/scores. 
This survey provides a systematic and comprehensive review of rank-based decomposable losses in machine learning. Specifically, we provide a new taxonomy of loss functions that follows the perspectives of aggregate loss and individual loss. We identify the \textit{aggregator} to form such losses, which are examples of set functions. We organize the rank-based decomposable losses into eight categories.
Following these categories, we review the literature on rank-based aggregate losses and rank-based individual losses. We describe general formulas for these losses and connect them with existing research topics.
We also suggest future research directions spanning unexplored, remaining, and emerging issues in rank-based decomposable losses.  
\end{abstract}

}

\maketitle

\IEEEdisplaynontitleabstractindextext

%
\IEEEpeerreviewmaketitle

\IEEEraisesectionheading{\section{Introduction}\label{sec:introduction}}

%
%
%
%
\IEEEPARstart{M}{achine} learning is instrumental to recent advances in artificial intelligence and big data analysis. They have been used in almost every area of computer science and many fields of natural sciences, engineering, and social sciences. Practical applications of machine learning algorithms are also abundant – it is not an exaggeration to say that without efficient and effective machine learning algorithms, many industries, such as internet advertisement, automatic driving, and social network mining, would not have flourished.
Most machine learning models are trained by minimizing a learning objective over a set of training samples. The learning objective comprises the learning loss corresponding to the errors on the training set, regularizers that control the model complexity, and constraints that incorporate conditions on the model parameters. 

In forming the learning objective, training loss is a critical component. The loss function is often constructed by aggregating a set of individual values into a single numerical value. The loss measures the quality of the model on a single training sample and is referred to as the \textit{individual loss}. For example, in binary classification, the individual loss usually quantifies the discrepancy between the prediction score and the label.
On the other hand, the loss that works over all training data is referred to as the \textit{aggregate loss}, which combines individual losses or scores (prediction scores from a model on samples) of a learning model over each training sample. For example, to evaluate the empirical risk of the model on a dataset, the aggregate loss uses the average of individual losses for all training examples. 
The prevalent practice in machine learning is first to choose the form of the individual loss, then construct the aggregate loss. 
For example, in binary classification, we first design the individual loss function based on the hinge function. Second, we calculate each individual loss of samples. Finally, we design aggregate loss based on the average operator to calculate the average loss score among all samples.  


The aggregate losses can be categorized as  decomposable and non-decomposable. The decomposable aggregate loss can be decomposed into individual terms for each training sample. The non-decomposable aggregate loss is a function of the entire set of training samples \cite{ranjbar2012optimizing}. For example, the most popular decomposable aggregate loss is the average loss, or empirical risk minimization \cite{vapnik1999nature}. This loss calculates the average value of all individual losses obtained from training samples. In addition, the non-decomposable aggregate losses such as the area under the ROC curve (AUC) loss \cite{yan2003optimizing} and the average precision (AP) loss \cite{baeza1999modern} are widely discussed in the topic of information retrieval. 
On the other hand, most existing individual losses can be decomposed into a combination of individual scores from each class label. 
For example, the cross-entropy loss \cite{murphy2022probabilistic} and the conventional multi-class loss \cite{crammer2001algorithmic}. In this survey, we focus on decomposable losses, including aggregate and individual losses.  We call decomposable aggregate losses as aggregate losses in the following whenever there is no confusion.

To form the decomposable aggregate losses or individual losses, the ranking order is the most basic relation among the individual values that needs to be considered.
\shu{Simply put, the ranking order is crucial as it helps determine the significance of each value, such as per-instance losses or per-label scores. By arranging these values in a specific order, we can prioritize the most important information and design better decomposable aggregate or individual loss functions that accurately reflect the desired outcomes. For instance, if we only have access to individual losses from a black-box model, we can use the ranking order of these losses to design a robust aggregate loss function that can improve the model's training. Thus, rank is linked to aggregate loss.
}

In addition, a set function \cite{kolmogorov1975introductory} needs to be applied to map a set of elements to a single value. Therefore, it is natural to define the rank-based set function, named \textit{aggregator}, to construct the specific decomposable aggregate loss and individual loss. For example, the aggregator works on a set of individual losses to construct a specific aggregate loss, and the aggregator works on a set of individual scores that can define an individual loss. Therefore, the rank-based decomposable losses are most important for designing machine learning models, and they have drawn much attention  \cite{fan2017learning, shen2019learning, shen2019iterative, shah2020choosing, roh2021sample, hu2020learning, hu2021sum, hu2021tkml, lapin2015top, lapin2016loss, lapin2017analysis} in the recent years. However, to the best of our knowledge, no systematic survey exists of the rank-based decomposable loss in the current machine learning literature. 


There are several important issues concerning rank-based decomposable losses.
\begin{enumerate}[leftmargin=*]
    \item \textbf{Definitions of rank-based decomposable aggregate losses and individual losses.} Finding a suitable method to classify and analyze the rank-based decomposable loss is significant. Through the existing works discussed in this survey, we categorize it into two aspects: aggregate loss from the sample level and individual loss from the label level. However, the formal definitions of these two notions are not evident in the current literature. 
    
    \item \textbf{Aggregator.} Since we need to design a rank-based decomposable loss, we must first determine a specific aggregator (or rank-based set function). The different aggregators may involve different meanings. However, there are no existing systematic studies about the aggregator. 
    In addition,  different aggregators may have relationships with each other. A loss designed based on one aggregator may extend to another task that only needs to change the current aggregator to another aggregator. Therefore, the flexibility of the aggregator is significant. However, a comprehensive relationship among these aggregators is unknown.  
    
    \item \textbf{Geometric  interpretability.}
    The interpretability of the different rank-based decomposable losses is vital in real-world applications because we need to provide convincing evidence to support the final obtained model performance. Several works \cite{blondel2020learning, lapin2015top, kong2020rankmax, lyu2020average} have explored the geometric interpretation of the designed rank-based decomposable losses. However, for more complicated rank-based decomposable losses, their interpretability is still not explored. How to define a general framework that can explain all rank-based decomposable losses in the ranking level is worth exploring.  
    
    \item \textbf{High training time complexity.}
    Training time complexity is significant for a learning model using a rank-based loss. Some aggregators usually involve sorting operations, which may cause more than $O(n\log n)$ time complexity, where $n$ is the total number of elements in a ranked list. Sorting can slow down training, especially for a large number of training samples or class labels. Finding an efficient optimization strategy is crucial for the applicability of rank-based decomposable losses.
    
\end{enumerate}
\vspace{-5mm}

\subsection{Related Works}
\vspace{-1mm}
There exist several surveys \cite{wang2021survey, jadon2020survey, tian2022recent, li2011short, ranjbar2012optimizing, li2014learning, yang2022auc, nie2018investigation, wang2020comprehensive} on loss functions. Specifically, \cite{wang2021survey, jadon2020survey, tian2022recent} focus on loss functions that are used in specific learning tasks or domains such as face recognition, semantic segmentation, computer vision, etc. 
\cite{nie2018investigation} investigates the commonly used loss functions in general machine learning, such as classification and regression. \cite{wang2020comprehensive} revisits many existing loss functions in machine learning from two aspects: traditional machine learning (classification, regression, and unsupervised learning) and deep learning (object detection and face recognition). However, the loss functions discussed in these surveys are only focused on the individual loss level. They assume the loss functions use the average operator in the sample level. On the other hand, 
\cite{li2011short, ranjbar2012optimizing, li2014learning, yang2022auc} discuss the losses from the information retrieval area. However, they only view the losses from the perspective of evaluation metrics. \textit{There is no system survey about rank-based decomposable Losses, especially view losses from the aggregator that aggregates a set of values to a single value. }
\vspace{-3mm}

\subsection{Contributions}
\vspace{-1mm}
\begin{enumerate}[leftmargin=*]
    \item This is the first comprehensive survey on rank-based decomposable losses in machine learning. Some of the relevant surveys in the past only focused on the loss functions from a specific topic. Others summarize very commonly used loss functions in machine learning. 
    Up to now, there is no dedicated and comprehensive survey on rank-based decomposable losses. This work fills this gap. Through this organized and systematic survey, we expect to help push research in this field forward.

    \item We organized this survey for rank-based decomposable loss functions from two novel aspects: aggregate loss level and individual loss level. This perspective has not been discovered in the past. Furthermore, this taxonomy can be extended to the general losses in machine learning, deep learning, data mining, etc., and can help  researchers to design loss more flexibly in the future.
    
    \item We introduce a new set function based on ranking, named aggregator, and use it to formulate aggregate and individual losses. Based on existing rank-based decomposable losses, we classify the aggregator into eight categories and analyze their relationships and properties.
 
    \item This survey reviews the most up-to-date rank-based decomposable losses in machine learning. The related topics include  general machine learning, deep learning, trustworthy machine learning, application, etc. Most of the papers are published in top-tier conferences and leading journals in machine learning, data science, deep learning, and artificial intelligence. We first summarize two types of loss functions based on our defined aggregator. Then, we comprehensively review existing works from the perspectives of rank-based aggregate loss and rank-based individual loss. In particular, each perspective is organized according to the types of aggregators. 
    
    \item 
    \shu{We also provide directions for future works, including new aggregators and nested aggregators, converting non-decomposable losses to decomposable ones, statistical machine learning theory, aggregate gradients,  hyperparameter learning, etc.} These opportunities indicate that the issues we discussed have not been fully solved.   
    
\end{enumerate}
The remainder of this survey is organized as follows. 
\begin{itemize}[leftmargin=*]
    \item In Section \ref{sec:background}, we introduce some notations and the formula of the learning objective.
    
    \item In Section \ref{sec:loss_functions}, \shu{we define rank-based and non-rank-based losses, decomposable and non-decomposable losses, and aggregate and individual losses. Additionally, we provide examples to help make these concepts clearer.}
    
    \item In Section \ref{sec:set_operation}, we introduce a rank-based set function named aggregator. We also define aggregate loss and individual loss based on the aggregator. Then we analyze the general properties of the aggregator.
    
    \item In Section \ref{sec:aggregate_loss}, we provide a collection of published works in rank-based aggregate losses and summarize them into different aggregators. 
    
    \item In Section \ref{sec:individual_loss}, we survey recent papers about rank-based individual losses and summarize them into different aggregators. 
    
    \item In Section \ref{sec:other_topics}, we connect the rank-based losses with several hot topics in machine learning to demonstrate their applicability and popularity.
    
    \item In Section \ref{sec:future_directions}, we discuss remaining and emerging issues in rank-based losses. In the meantime, we provide suggestions about the directions for future works. We conclude this survey in Section \ref{sec:conclusion}. 
\end{itemize}
\vspace{-3mm}

\section{Background}\label{sec:background}

\subsection{Notation} \label{sec:notation}
\vspace{-1mm}
We introduce some necessary notations that will be used in the following sections. Let $\mathbb{R}$ be the real domain and $\mathbb{R}_+$ is the nonnegative domain of $\mathbb{R}$. Denote $\mathbb{N}_n$ as the set $\{1,\cdots,n\}$, $\mathbb{N}_C$ as the set $\{1,\cdots,C\}$, and $\mathbb{I}_{a}$ as an indicator function with $\mathbb{I}_{a}=1$ if $a$ is true and 0 otherwise. 
A hinge function is defined as $[x]_+=\max\{0,x\}$. Let $\|\x\|_1$, $\|\x\|_2$, $\|\x\|_\infty$ be the $l_1$, $l_2$, and $l_\infty$ norms of a vector $\x$, respectively. For a set $S=\{s_1,\cdots,s_n\}\subset \mathbb{R}^n$, we define its ranked version as an ordered set (ranked list) $\left<S\right>=\left<s_{[1]},\cdots,s_{[n]}\right>$, with $s_{[1]}\geq\cdots\geq s_{[n]}$, obtained by sorting elements of $S$ in the descending order. $|S|$ is the cardinality of a set $S$. $s_{[k]}$ denotes the top-$k$ element, which is the $k$-th largest element in $S$ (as well as $\left<S\right>$). Without loss of generality, we only consider $S$ with no ties since ties can be broken in any consistent way.
\vspace{-3mm}

\subsection{Learning Objective}
\vspace{-1mm}
The central task of a machine learning algorithm is to ``train'' a model using training data, which entails seeking models that minimize certain performance metrics known as the \textit{losses}. The losses are usually combined with the regularization terms and constraints to form a \textit{learning objective}, which is a quantity to be optimized in training. 
A general learning objective can be represented as follows,
\vspace{-1mm}
\begin{equation*}
    \begin{aligned}
    \min_f \ \mathcal{L}(f, \mathcal{D}) + \alpha \Omega(f), \ \ \ 
    \mbox{s.t.} \ \mathcal{C},
    \end{aligned}
\vspace{-1mm}
\end{equation*}
where $f\in \mathcal{H}$ is a function (\ie, learning model.) from the hypothesis space (function family) $\mathcal{H}$. $\mathcal{D}$ is a training dataset. $\mathcal{L}(\cdot, \cdot)$ is the loss or empirical risk. $\Omega(f)$ is the regularization term representing the model complexity. $\alpha$ is the trade-off to balance the loss and the model complexity. $\mathcal{C}$ represents the constraints for specific learning scenarios. The loss is the core part of the learning objective. 
\vspace{-3mm}

\begin{figure}[t]
\vspace{-\intextsep}
\centering
\includegraphics[trim=1 1 1 1, clip,keepaspectratio, width=0.45\textwidth]{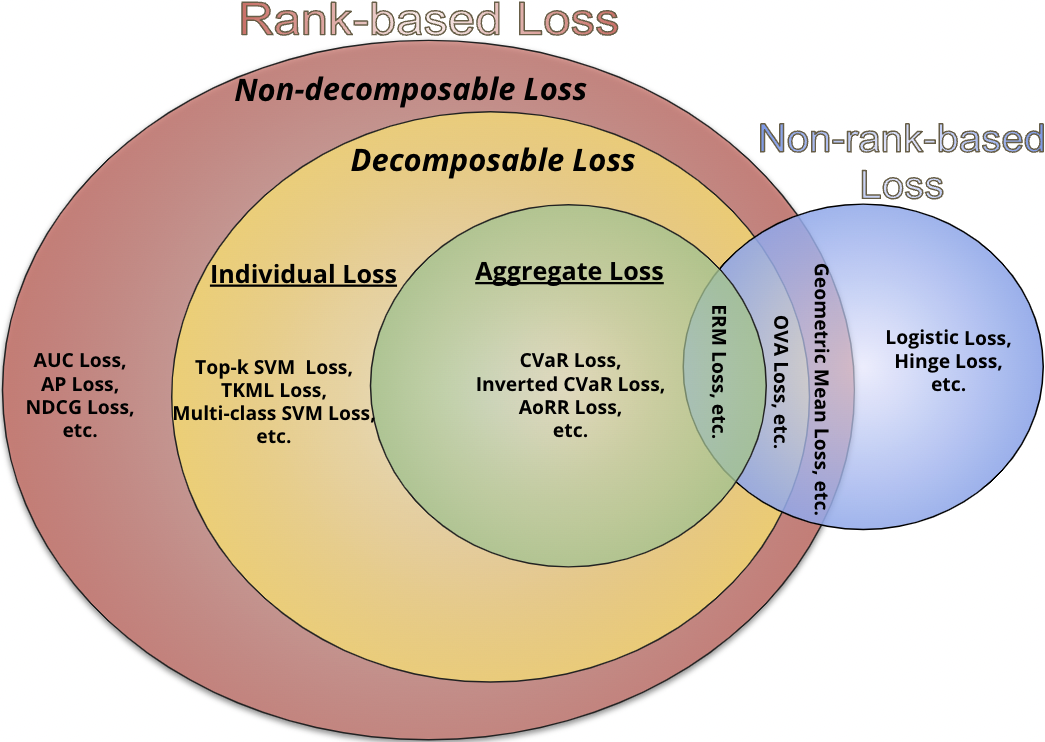}
\vspace{-0.2cm}
\caption{\it \shu{A Venn diagram illustrates the relationships among rank-based, non-rank-based, decomposable, non-decomposable, aggregate , and individual losses.} 
}
\vspace{-0.5cm}
\label{fig:Venn-diagram}
\end{figure}

\section{Loss Functions}\label{sec:loss_functions}
\vspace{-1mm}
\shu{
In this section, we introduce the definitions of rank-based and non-rank-based losses, decomposable and non-decomposable losses under rank-based losses,  and aggregate and individual losses. The relationships of these losses are illustrated in Fig. \ref{fig:Venn-diagram}. We also give some examples to help clarify these concepts. 
}
\vspace{-3mm}


\subsection{Rank-based Loss and Non-rank-based Loss}
\vspace{-1mm}
\shu{From a ranking perspective, the losses in machine learning can be summarized into rank-based and non-rank-based losses. They can be defined as follows:}

\begin{definition}
\shu{(Rank-based loss). The rank-based loss is a function designed in a way that is \textit{aware} of the order relevance of the items, such as prediction scores assigned by a model to possible labels or classes  for a given instance or loss values that measure how well a model is performing on given instances.} 
\end{definition}

\begin{definition}
\shu{(Non-rank-based loss). The non-rank-based loss is a function designed in a way that is \textit{agnostic} of the order relevance of the items.} 
\end{definition}

\shu{Rank-based losses have received significant attention in machine learning, with many losses, such as AUC loss \cite{yan2003optimizing}, AP loss \cite{baeza1999modern}, Average of Ranked Range (AoRR) loss \cite{hu2020learning}, and Top-$k$ Multi-Label (TKML) loss \cite{hu2020learning}, being designed based on the relative order of items. Non-rank-based losses are also popular, such as logistic loss \cite{kleinbaum2002logistic}, hinge loss \cite{cortes1995support}, and negative variance loss \cite{wang2020comprehensive}. }

\shu{\textbf{Overlaps}.
Rank-based and non-rank-based losses are not mutually exclusive, and there are overlaps between them as shown in Fig. \ref{fig:Venn-diagram}. For example, the Empirical Risk Minimization (ERM) \cite{vapnik1999nature} approach in machine learning involves minimizing the expected risk of a model by minimizing the empirical risk over the training data, which is typically calculated by aggregating the loss values of individual instances using a specific loss function, such as the logistic loss. In this case, the loss function does not need to consider the order relevance of items and is therefore a non-rank-based loss. However, when information about the ranking or relative importance of loss values is available, we can directly incorporate this information into the ERM loss function, making it a rank-based loss. Similarly, One-versus-All (OVA)\cite{vapnik1999nature}, geometric mean\cite{ando2004geometric, su2017learning}, and harmonic mean\cite{wang2021harmonic} losses can also belong to both categories. For completeness, we consider average and sum operator-controlled losses as rank-based losses in the following discussion.}  

\shu{\textbf{Importance}.
We focus on rank-based losses in this survey since they are crucial in machine learning. For example, for ranking problems, they directly optimize the ranking performance of a model by explicitly modeling the ranking order of items. This is important in applications like information retrieval \cite{liu2009learning}, search engines, and recommender systems \cite{ortis2019predicting}, where the goal is to return the most relevant items to the user. Additionally, rank-based losses are more robust to noise and outliers in the data compared to non-rank-based losses, which can be sensitive to significant errors in the predicted score or particular instance's loss value \cite{hu2020learning}. Therefore, rank-based losses offer advantages over non-rank-based losses, including flexibility and robustness.}
\vspace{-3mm}

\subsection{Decomposable Loss and Non-decomposable Loss}\label{sec:Decomposable-Non-decomposable-Loss}
\vspace{-1mm}
\shu{We categorize rank-based losses as decomposable or non-decomposable. We explain these categories using the sample level loss (\ie, loss defined on all instances in the training set) as described in \cite{ranjbar2012optimizing}. However, their definitions can also apply to label level loss (\ie, loss defined on prediction scores of possible labels or classes for a given instance). Let $\mathcal{X}$ and $\mathcal{Y}$ be the input feature domain and target domain, respectively, and let $\mathcal{Z}=\mathcal{X}\times\mathcal{Y}$ be the joint domain. 
The training dataset is denoted as $\mathcal{D}:=\{\z_1,\cdots,\z_n\}$, where each $\z_i=(\x_i,\y_i)$ is a finite subset of $\mathcal{Z}$. The goal of most learning problems is to find a function $f\in \mathcal{H}$ that optimizes the expected prediction performance on a new dataset $\mathcal{D}':=\{\z'_1,\cdots,\z'_{n'}\}$, where each $\z'_i=(\x'_i,\y'_i)$. This can be achieved by minimizing the risk function $\mathcal{R}(f)$: 
\vspace{-1mm}
\begin{equation*}
    \mathcal{R}(f)= \int \Gamma\big([f(\x'_1), ...,f(\x'_{n'})], [\y'_1,...,\y'_{n'}]\big) d \mathbb{P}(\mathcal{D}'),
\vspace{-1mm}
\end{equation*}
where $\Gamma$ is a loss function and $\mathbb{P}(\mathcal{D}')$ is the probability distribution on $\mathcal{D}'$.} 

\shu{\textbf{Decomposable loss}. It can be defined as:}
\begin{definition}
\shu{(Decomposable loss). A loss that \textit{can} be decomposed into individual losses or scores for each instance or label is called a decomposable loss. Each individual loss or score can be calculated independently of the others.}
\end{definition}

\shu{Many algorithms, such as support vector machine (SVM) \cite{cortes1995support}, assume that the instances are i.i.d. and the loss function $\Gamma$ is decomposable. In this case, $\Gamma$ can be decomposed into a linear combination of a loss function $\ell$ over instances. As a result, the decomposable risk function $\mathcal{R}_{\text{dec}}(f)$ can be expressed as $\mathcal{R}(f)=\mathcal{R}_{\text{dec}}(f)= \int \ell\big(f(\x'), \y'\big) d \mathbb{P}(\x', \y')$.
Rather than minimizing the estimated risk function, $\mathcal{R}_{\text{dec}}(f)$, learning algorithms approximate it with the empirical risk or loss function $\mathcal{L}(f, \mathcal{D})$, which is defined as: 
$\mathcal{L}(f, \mathcal{D}) =\frac{1}{n}\sum_{i=1}^n \ell(f(\x_i),\y_i)$. 
As an illustration, we use the average operator, but in the subsequent sections, we will showcase other decomposable losses based on ranking.}

\shu{\textbf{Non-decomposable loss}. It can be defined as: }
\begin{definition}
\shu{(Non-decomposable loss). A loss that \textit{cannot} be decomposed into {per-instance losses or per-label scores}, and instead requires a joint optimization over all instances or labels is called a non-decomposable loss. }
\end{definition}
\shu{In the non-decomposable loss, $\Gamma$ cannot be decomposed. 
Therefore, we need to find an algorithm that can directly optimize the empirical risk based on the following loss:
\vspace{-1mm}
\begin{equation*}
    \mathcal{L}_{\text{non-dec}}(f, \mathcal{D}) =\Gamma\big((f(\x_1), ...,f(\x_{n'})), (\y_1,...,\y_{n'})\big).
\vspace{-1mm}
\end{equation*}
Optimizing $\mathcal{L}_{\text{non-dec}}(f, \mathcal{D})$ for an arbitrary loss function $\Gamma$ can be computationally challenging since such a loss function usually only partially or does not decompose over samples. 
A prominent example is the area under ROC curve (AUC) loss \cite{yan2003optimizing}, which is defined on the prediction scores of each training sample $\x_i$, $f(\x_i)$, and is equivalent to the Wilcoxon-Mann-Whitney statistic \cite{hanley1982meaning} as $\frac{1}{|\mathcal{I}^+||\mathcal{I}^-|}\sum_{i\in \mathcal{I}^+}\sum_{j\in \mathcal{I}^-} (\mathbb{I}_{f(\x_i)<f(\x_j)}+\frac{1}{2}\mathbb{I}_{f(\x_i)=f(\x_j)})$, where we consider binary classification problem and denote $\mathcal{I}^+=\{i|\y_i=+1\}$ and $\mathcal{I}^-=\{i|\y_i=-1\}$ as the sets of indices of positive and negative samples, respectively. However, the discontinuous indicator function in the AUC loss makes its direct minimization an NP-hard problem \cite{hand2001simple}. As such, most existing AUC learning algorithms (\eg, \cite{zhao2011online,gao2013one}) replace the indicator function with surrogates that are continuous and convex upper-bounds of the AUC. Specifically, the surrogate loss function takes the form as $\frac{1}{|\mathcal{I}^+||\mathcal{I}^-|}\sum_{i\in \mathcal{I}^+}\sum_{j\in \mathcal{I}^-} \ell'(f(\x_j)-f(\x_i))$, and common choices for $\ell'$ include the pairwise hinge loss \cite{gao2013one}, $\ell'_h=[1+(f(\x_j)-f(\x_i))]_+$,
and the pairwise logistic loss \cite{sulam2017maximizing}, $\ell'_{lg}=\log_2(1+e^{f(\x_j)-f(\x_i)})$. 
The AUC loss and its surrogates are examples of non-decomposable losses that cannot be decomposed into a summation of individual terms over all training samples. Other examples include (Normalized) Discounted Cumulative Gain ((N)DCG) \cite{jarvelin2017ir} and Recall loss \cite{rolinek2020optimizing}, as well as their discussions, can be found in \cite{liu2009learning, li2011short}.}

\shu{\textbf{Difference}. The major difference between these two types of losses is that decomposable losses can be decomposed into per-instance losses or per-label scores, while non-decomposable losses cannot. 
Optimizing decomposable losses is computationally efficient since it can be done independently for each instance or label. In contrast, optimizing non-decomposable losses requires joint optimization over all instances or labels and often requires specialized optimization methods.
For example, computing a single gradient update for a non-decomposable loss generally has a quadratic cost in the number of training examples.  
So it is often necessary to develop methods that convert or bound non-decomposable losses with decomposable ones.}
\vspace{-3mm}

\begin{figure*}[t]
\centering
\includegraphics[trim=1 1 1 1, clip,keepaspectratio, width=1\textwidth]{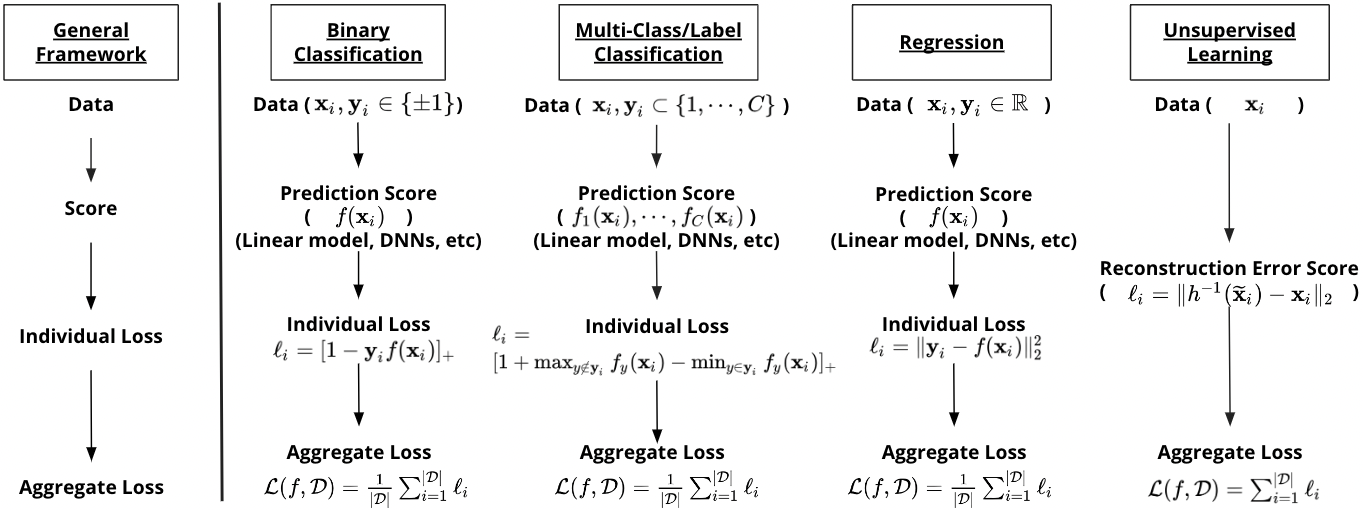}
\vspace{-0.6cm}
\caption{\it Illustrative examples of the relationships between individual loss and aggregate loss in binary classification, multi-class/label classification, regression, and unsupervised learning.
}
\vspace{-0.5cm}
\label{fig:losses}
\end{figure*}

\subsection{Aggregate Loss and Individual Loss }
\vspace{-1mm}
We refer to the loss over all training data as the \textbf{aggregate loss}, in order to distinguish it from the
\textbf{individual loss} that measures the quality of the model on a single training sample. The relationship between aggregate loss and individual loss can be found in Fig. \ref{fig:losses}. 
They can be expressed as,
\vspace{-1mm}
\begin{equation*}
    \begin{aligned}
    \mathcal{L}(f, \mathcal{D}) = \underbrace{\mathcal{L}\big(\{\overbrace{\ell(f(\x_i),\y_i)}^{\textbf{\mbox{individual loss}}}|i\in \mathbb{N}_n\}\big)}_{\textbf{\mbox{aggregate loss}}},
    \end{aligned}
\vspace{-1mm}
\end{equation*}
where the aggregate loss function $\mathcal{L}(\cdot)$ can be constructed by a linear combination of individual loss function $\ell(\cdot,\cdot)$ over individual examples. For example, the most popular decomposable aggregate loss is the average loss, which is also called empirical risk \cite{vapnik1999nature}.

\shu{An aggregate loss can be either \textit{decomposable} or \textit{non-decomposable}. 
The decomposable aggregate loss is formed by individual losses, which can be calculated independently for each instance, while the non-decomposable aggregate loss cannot. An example of a non-decomposable aggregate loss is the AUC loss, as discussed in Section \ref{sec:Decomposable-Non-decomposable-Loss}.  
However, this survey focuses on decomposable aggregate losses since non-decomposable losses have already been widely studied in the literature, particularly in the field of information retrieval  \cite{liu2009learning, li2011short}. We refer to decomposable aggregate loss as just aggregate loss. Below are some examples of how aggregate and individual losses are utilized in both supervised and unsupervised learning.}



\textbf{Supervised learning}. For a supervised learning problem, the goal of the learning task is to find a parametric predictor (or model) $f:\mathcal{X}\rightarrow \mathcal{Y}$ that can be used to predict the label or value of a new sample $\x$. We introduce an \textit{individual loss} $\ell$ as a bi-variate function $\ell:\mathcal{Y}\times\mathcal{Y}\rightarrow \mathbb{R}_+$ for evaluating the fitting quality of $f$. $\ell(f(\x), \y)$ quantifies the discrepancy between the prediction $f(\x)$ and ground truth target $\y$, with $\ell(f(\x),\y)=0$ when they are in agreement. 
\begin{itemize}[leftmargin=*]
    \item The explicit form of individual loss is dominated by a specific learning task. 
    \begin{itemize}[leftmargin=*]
        \item In binary classification (\ie, $\y\in\mathcal{Y}=\{\pm 1\}$), $\ell$ can be the 0/1 loss $\mathbb{I}_{\y f(\x)<0}$, the logistic loss $\log_2(1+e^{-\y f(\x)})$, the hinge loss $[ 1-\y f(\x)]_+$, and the binary cross entropy loss $-\y\log f(\x)-(1-\y)\log(1-f(\x))$, etc.
        \item In multi-class classification (\ie, $\y\in\mathcal{Y}=\{1, \cdots, C\}$, where $C> 2$.), $\ell$ can be the conventional multi-class loss \cite{crammer2001algorithmic}: $\max_{j\in\mathcal{Y}}\{\mathbb{I}_{j\neq\y}+f_j(\x)-f_{\y}(\x)\}$, where $f_j(\x)\in \mathbb{R}$ corresponding to the prediction score of $\x$ with regards to the $j$-th class. 
        \item In multi-label classification (\ie, $\y\subset\mathcal{Y}=\{1, \cdots, C\}$, where $C> 2$.), $\ell$ can be the conventional multi-label loss \cite{crammer2003family}: $[1+\max_{j\notin \y} f_j(\x)-\min_{j\in\y}f_j(\x)]_+$.
        \item In regression (\ie, $\y\in\mathcal{Y}=\mathbb{R}$), $\ell$ can be the squared $l_2$ difference $\|\y-f(\x)\|_2^2$. 
    \end{itemize}
    More individual losses can be found in  \cite{wang2020comprehensive}.
    \item The most popular aggregate loss is the average loss \cite{vapnik1999nature}, which using the average of individual losses for all training samples and can be defined as $\mathcal{L}(f, \mathcal{D})=\frac{1}{n}\sum_{i\in \mathbb{N}_n}\ell(f(\x_i),\y_i)$. The maximum aggregate loss \cite{shalev2016minimizing} $\mathcal{L}(f, \mathcal{D})=\max_{i\in \mathbb{N}_n}\ell(f(\x_i),\y_i)$ is also widely used in the supervised learning tasks. It uses the maximum individual loss to learn the model.
\end{itemize}


\textbf{Unsupervised learning}. In unsupervised learning, there are two major learning tasks: clustering and dimension reduction. For clustering, the objective is to divide samples into different clusters according to the similarity index. For dimension reduction, a projector is applied to project the original high-dimensional feature space into low-dimensional space. Then the similarity is calculated between these two spaces to determine the preserving ability of data structure and usefulness. Since there are no ground truth targets in the learning problems, the individual loss will only depend on the scores and features. 
For example,
\begin{itemize}[leftmargin=*]
    \item The implicit form of individual loss can be the reconstruction error in dimension reduction problems, which is applied to measure the distance between the original sample and the reconstructed sample according to the inverse projection mapping. For example, in principal component analysis (PCA) \cite{wold1987principal},  the loss can be defined as $\|h^{-1}(\widetilde{\x}_i)-\x_i\|_2$, where $\widetilde{\x}_i$ is the low-dimensional representation of $\x_i$ and $h$ is the projection mapping. 
    \item The aggregate loss can be an average or sum of reconstruction errors over all samples, as  $\sum_{i=1}^n\|h^{-1}(\widetilde{\x}_i)-\x_i\|_2$.

\end{itemize}
\vspace{-3mm}


\section{Aggregator for Decomposable Losses}\label{sec:set_operation}
\vspace{-1mm}
In this section, we will introduce and define a rank-based set function, named aggregator. Then we will discuss its general properties.
\vspace{-3mm}


\subsection{Aggregator}\label{sec:aggregator}
\vspace{-1mm}
For a set of real numbers representing individual values, the ranking order reflects the most fundamental relation among them. Therefore, designing aggregate loss and individual loss can be achieved by choosing operations defined based on the ranking order of the individual values. Such individual values can be the individual losses for aggregate loss on the data sample level or can be the individual scores for individual loss on the data label level, where the number of labels should be greater than 1. 
Therefore, in this survey, we focus on rank-based aggregate loss and rank-based individual loss.


At a more general level, we noticed that once the set of all individual losses, $\ell(\mathcal{D}):=\{\ell_1,\cdots, \ell_n\}$ with $\ell_i=\ell(f(\x_i), \y_i)$ (or $\ell_i=\ell(f(\x_i))$ if targets (labels) are not available in unsupervised learning scenario), is determined, a decomposable aggregate loss can be constructed without concerning the form of the learning model $f$, the choice of the individual loss $\ell$, and the training data $\mathcal{D}$. 
\shu{Instead, we only need to determine which individual losses from the set $\ell(\mathcal{D})$ should be extracted for aggregation, and their order of importance can be indicated by ranking.
This means that a decomposable aggregate loss can be formulated by a rank-based \textit{set function}} \cite{kolmogorov1975introductory} that maps a set, \ie, $\ell(\mathcal{D})$, to a real number while considering the ranking order among elements in the set. In this survey, we term such function as an \textbf{\textit{aggregator}}. Therefore, the relationship of aggregate loss, aggregator, and individual loss is as follows,
\vspace{-1mm}
\begin{equation*}
    \begin{aligned}
    \underbrace{\mathcal{L}(\ell(\mathcal{D}))}_{\textbf{\mbox{aggregate  loss}}}:=\underbrace{\mathcal{L}}_{\textbf{\mbox{aggregator}}}\big(\{\underbrace{\ell_i}_{\textbf{\mbox{individual  loss}}}|i\in \mathbb{N}_n\}\big).
    \end{aligned}
\vspace{-1mm}
\end{equation*}
Note that $\mathcal{L}(\ell(\mathcal{D}))$ is just  $\mathcal{L}(f,\mathcal{D})$, which we used in the previous section. For instance, the average loss is constructed from the average aggregator that computes the arithmetic average of a set $\mathcal{L}_{avg}(\ell(\mathcal{D}))=\frac{1}{n}\sum_{i=1}^n\ell_i$. The maximum loss corresponds to the maximum aggregator, $\mathcal{L}_{max}(\ell(\mathcal{D}))=\max_{i\in \mathbb{N}_n}\ell_i$. Similarly, the average top-$k$ (AT$_k$) loss originates from the AT$_k$ aggregator, which is defined as $\mathcal{L}_{at-k}(\ell(\mathcal{D}))=\frac{1}{k}\sum_{i=1}^k\ell_{[i]}$.


In the aggregate loss, the aggregator works at the data sample level. However, the aggregator can also work at the label level, which corresponds to the aggregator for the individual loss. Therefore, the relationship between  individual loss, aggregator, and individual score is as follows,
\vspace{-1mm}
\begin{equation*}
    \begin{aligned}
    \underbrace{\ell_i}_{\textbf{\mbox{individual  loss}}}:=\underbrace{\mathcal{L}}_{\textbf{\mbox{aggregator}}}\big(\{\mathcal{O}(\underbrace{f_j(\x_i)}_{\textbf{\mbox{individual  score}}}, \y_i)|j\in \mathbb{N}_C\}\big),
    \end{aligned}
\vspace{-1mm}
\end{equation*}
where $f_j(\x_i)$ is the prediction score of $\x_i$ on $j$-th class. $\mathcal{O}$ is a function, which can output a value based on $f_j(\x_i)$ and $\y_i$. To simplify the notation, we let $\mathbb{O}(f(\x_i))=\{\mathcal{O}(f_j(\x_i), \y_i)|j\in \mathbb{N}_C\}$. The rank-based individual losses are widely used in multi-label and multi-class learning. 
For example, let us assume a general multi-label classification problem with a total of $C> 2$ possible labels. For an input $\x_i\in \mathcal{X}$, its true labels are represented by a set of labels $\oldemptyset \neq \y_i\subset\{1,\cdots, C\}$. We introduce a continuous multi-label prediction set $F(\x_i)=\{f_1(\x_i),\cdots, f_C(\x_i)\}$ and $\mathcal{O}(f_j(\x_i),\y_i)=[1+f_j(\x_i)\times \mathbb{I}_{j\notin \y_i}-\min_{y\in \y_i}f_y(\x_i)]_+$. The conventional multi-label loss \cite{crammer2003family} can be constructed from the maximum aggregator such as $\ell_i=\mathcal{L}_{max}(\mathbb{O}(f(\x_i)))= \max_{j\in\mathbb{N}_C}\mathcal{O}(f_j(\x_i),\y_i)=[1+\max_{j\notin \y_i}f_j(\x_i)-\min_{j\in \y_i}f_j(\x_i)]_+ $. Note that it becomes the conventional multi-class loss \cite{crammer2001algorithmic} when $|\y_i|=1, \forall i\in \mathbb{N}_n$. 

\vspace{-3mm}

\subsection{General Properties of Aggregator} \label{sec:prop_agg}
\begin{figure}[t]
\vspace{-\intextsep}
\centering
\includegraphics[trim=1 1 1 1, clip,keepaspectratio, width=0.45\textwidth]{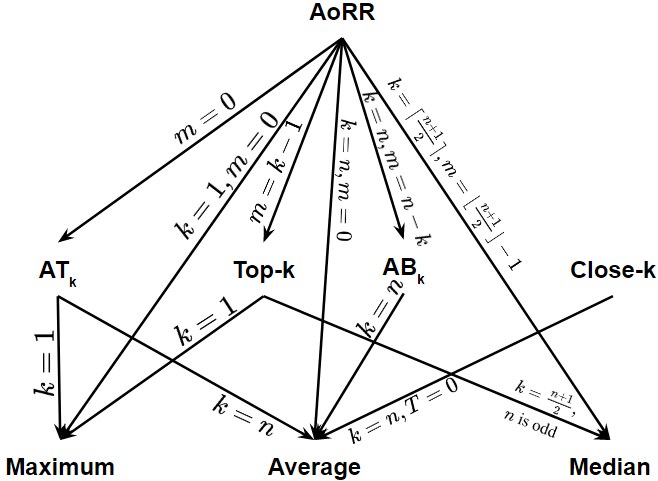}
\vspace{-0.2cm}
\caption{\it The relationships between aggregators. $k$ or $m$ are the hyperparameters of the source aggregator. $n$ is the total number of samples (for aggregate loss) or total number of labels (for individual loss). And $0\leq m< k\leq n$. $T$ is a specific hyperparameter for Close-k aggregator. 
}
\vspace{-0.5cm}
\label{fig:aggregators}
\end{figure}

\vspace{-1mm}
Before we study the properties of the aggregator, we first summarize all existing aggregators in the literature. We will introduce more details about using them in the following sections. By using the notation from Section \ref{sec:notation}, the existing aggregators can be organized as follows, 
\begin{enumerate}[leftmargin=*]
    \item \textbf{Average}:

    \centerline{$\mathcal{L}_{avg}(S)=\frac{1}{n}\sum_{i=1}^ns_i.$}
    
    Note that there is a variant of the average aggregator, which omits the factor $\frac{1}{n}$ and then becomes the sum aggregator. We do not distinguish `sum' from `average' for notational clarity.  This also holds for the following aggregators. 
    \item \textbf{Maximum}:

    \centerline{$\mathcal{L}_{max}(S)=\max_{1\leq i\leq n}s_i=s_{[1]}.$}
    
    
    \item \textbf{Median}: 

    \centerline{$\mathcal{L}_{med}(S)={1 \over 2}
    \left(s_{\left[\lfloor \frac{n+1}{2}\rfloor\right]} + s_{\left[\lceil \frac{n+1}{2} \rceil\right]}\right).$}
    

    \item \textbf{Top-$k$}:

    \centerline{$\mathcal{L}_{top-k}(S)=s_{[k]}.$}
    
    
    \item \textbf{Average Top-$k$ (AT$_k$)}:

    \centerline{$\mathcal{L}_{at-k}(S)=\frac{1}{k}\sum_{i=1}^ks_{[i]}.$}
    
    
    \item \textbf{Average Bottom-$k$ (AB$_k$)}:
    
    \centerline{$\mathcal{L}_{ab-k}(S)=\frac{1}{k}\sum_{i=n-k+1}^ns_{[i]}$.}
    
    \item \textbf{Average of Ranked Range (AoRR)}: 
    
    \centerline{$\mathcal{L}_{aorr}(S)=\frac{1}{k-m}\sum_{i=m+1}^ks_{[i]}$.}
    
    \item \textbf{Close-$k$}: 
    
    \centerline{$\mathcal{L}_{close-k}(S)=\sum_{i=1}^n\Big[s_{[n-i+1]}\cdot\mathbb{I}_{i\leq k}+M\cdot\mathbb{I}_{i>k \& \text{condition}}\Big]$,}    
    
    where `condition' means $s_{[n-i+1]}$ incorrectly  classified, $M$ is a preset constant, and $s_{[n-i+1]}$ is the individual loss $s_j$ with the ($n-i+1$)-th largest $|s_j-T|$ such that T is another preset constant.
\end{enumerate}
Fig.\ref{fig:aggregators} illustrates the relationships between these aggregators.

By further studying the \shu{characteristics} of general aggregator, we can gain more insights into its role in machine learning algorithms. 
We identify a list of \shu{characteristics} that desirable aggregators tend to possess: 

\begin{enumerate}[leftmargin=*]
    \item \textbf{Invariant to permutations}: $\mathcal{L}(S)$ should not change as the elements of $S$ undergone a permutation operation, \ie, $\mathcal{L}(S)=\mathcal{L}(\left<S\right>)$.
    \item \textbf{Stable with input size}: a large set does not have an advantage over a smaller set as input,  \ie, $\mathcal{L}(S)/|S|=O(1)$.
    \item \textbf{(Convexity,) continuous and differentiable}: an aggregator invariant to permutation is uniquely determined by a function of the vector with $s_1\geq\cdots \geq s_n$ and that function is differentiable (and convex \cite{boyd2004convex}).
\end{enumerate}

For instance, in the aggregate loss scenario, it is not difficult to prove that $\mathcal{L}_{avg}$ satisfies \shu{characteristics} 1)-3). In \cite{fan2017learning}, the authors show that the AT$_k$ aggregator is permutation invariant, size invariant, convex and continuous, and so does $\mathcal{L}_{max}$ as a special case of $\mathcal{L}_{at-k}$. On the other hand, the top-$k$ aggregator, $\mathcal{L}_{top-k}(S)=s_{[k]}$, which leads to the top-$k$ loss, \ie, $\mathcal{L}_{top-k}(\ell(\mathcal{D}))=\ell_{[k]}$ as the $k$-th largest individual loss, is a continuous but nonconvex and non-differentiable aggregator. This is because $s_{[k]}=k\mathcal{L}_{at-k}(S)-(k-1)\mathcal{L}_{at-(k-1)}(S)$. Therefore, it is the difference of two convex functions. In general, it is not a convex function of sorted elements of $S$. 
\shu{Furthermore, the aggregators corresponding to the geometric mean and harmonic mean, which are defined as $\mathcal{L}_{geo}(S)=(\prod_{i=1}^n s_i)^{\frac{1}{n}}$ and $\mathcal{L}_{har}(S)=(\sum_{i=1}^n s_i^{-1})^{-1}$, respectively, are continuous, differentiable but nonconvex aggregators. However, all of them are for non-decomposable losses. Hence, we do not include them in this survey.}


As we discussed before, for a set of real numbers representing individual values, the ranking order reflects the most basic relation among them. Therefore, designing aggregate losses and individual losses can be achieved by choosing operations of aggregator defined based on the ranking order of the individual values. This makes the top-$k$ aggregator $\mathcal{L}_{top-k}$ the basic building block of permutation invariant aggregators. A special case of $\mathcal{L}_{top-k}$ is when $k=\lfloor \frac{n+1}{2}\rfloor$ (assume $n$ is odd), which corresponds to the median aggregator that is known to be robust to outliers in individual losses. However, as shown earlier, one drawback of $\mathcal{L}_{top-k}$ is that it is nonconvex, which makes the aggregate loss difficult to use in practice. On the other hand, there are some interesting relations between $\mathcal{L}_{top-k}$ and $\mathcal{L}_{at-k}$: both are generalizations of the maximum operator $\mathcal{L}_{max}$ ($k=1$), and the latter is a tight convex upper-bound of the former, as $\mathcal{L}_{at-k}\geq \mathcal{L}_{top-k}$ with equality holds when all $s_i$ are the same. All these suggest that the AT$_k$ aggregator has some uniqueness in all rank-based aggregators. $\mathcal{L}_{at-k}$ can be the minimal convex rank-based aggregator, and $\mathcal{L}_{at-k}$ of different degree $k$ form a subspace of rank-based aggregator that can be used to approximate other aggregators based on order statistics \cite{arnold2008first}.
\vspace{-3mm}


\section{Rank-based Aggregate Losses }\label{sec:aggregate_loss}
\vspace{-1mm}
In this section, we will introduce seven types of rank-based aggregate losses and connect them with many popular topics in machine learning.
\vspace{-3mm}


\subsection{Average Aggregate Loss}
\vspace{-1mm}
The prevalent practice in machine learning is first to choose the form of the individual loss, then construct the aggregate loss using the average of individual loss for all training examples. We term this specific aggregate loss as the average (aggregate) loss \cite{vapnik1999nature}, which can be formulated by
\vspace{-1mm}
\begin{equation*}
    \begin{aligned}
    \mathcal{L}_{avg}(\ell(\mathcal{D}))=\frac{1}{n}\sum_{i=1}^n\ell_i=\frac{1}{n}\sum_{i=1}^n \ell_{[i]},
    \end{aligned}
    \vspace{-1mm}
\end{equation*}
where $\ell_i=\ell(f(\x_i), y_i)$ and $\ell_{[i]}$ is the top-$i$ individual loss in the set $\{\ell_1,\cdots, \ell_n\}$. From the above, the average loss can also be viewed as a rank-based aggregate loss.

The average loss is the dominant  paradigm for performance evaluation in machine learning. A possible consensus \cite{holland2021designing} is that it comes from the Perceptron of Frank Rosenblatt in the 1950s \cite{rosenblatt1958perceptron}, and the proof in the 1962s of the convergence of the perceptron learning algorithm for linearly separable data \cite{novikoff1963convergence} \cite[Chapter 11]{minsky1969introduction}. This has led to the development of  \textit{statistical learning theory} \cite{vapnik1999nature}.
The average loss is favored for several reasons, including its simplicity and the ease of finding individual loss functions relevant to the task. For example, logistic loss, hinge loss, or cross-entropy loss are common choices for classification tasks, while squared $l_2$ loss, absolute loss, or Huber loss are typical for regression. Additionally, the average aggregate loss can be interpreted as \textit{empirical risk minimization} (ERM) \cite{bartlett2006convexity, de2005model, steinwart2003optimal, wu2006learning} in the supervised learning or \textit{maximum (log)-likelihood} in unsupervised learning \cite{friedman2001elements}, providing a theoretical basis for this approach. 
Finally, stochastic gradient descent \cite{robbins1951stochastic} methods can be used to minimize the average loss on large datasets.
\shu{Several typical loss functions are shown in Table \ref{table:average}.}

\smallskip
\noindent
\shu{\textbf{Discussion}. The average aggregate loss is a popular and uncomplicated loss function used in many tasks such as regression, classification, and clustering. It has benefits such as being easy to compute, interpret, and optimize. However, it treats all individual losses equally, regardless of their significance, which makes it unsuitable for tasks where some samples are more important or when the data is imbalanced.}

\vspace{-1mm}

\begin{table}[h]
\centering
\scalebox{0.94}{
\begin{tabular}{l|l|l}
\hline
Work & Connection & Form \\ \hline
\cite{friedman2001elements}& Unsupervised& $\sum_{i=1}^n -\log f(\x_i)$ \\ \hline
\cite{bartlett2006convexity}& Supervised (Classification)& $\frac{1}{n}\sum_{i=1}^n([1-y_if(\x_i)]_+)^2$ \\ \hline
\cite{de2005model}  & Supervised (Regression) & $\frac{1}{n}\sum_{i=1}^n(y_i-f(\x_i))^2$ \\ \hline
\end{tabular}
}
\vspace{-0.5em}
\caption{\em \shu{Typical loss functions of the average aggregate loss.}
}
\label{table:average}
\vspace{-0.5cm}
\end{table}

\vspace{-3mm}

\subsection{Maximum Aggregate Loss}
\vspace{-1mm}
The average aggregate loss has limitations in dealing with imbalanced data distributions, as highlighted in \cite{shalev2016minimizing}. The paper explores alternative choices for aggregate loss, such as the maximum (aggregate) loss.
\vspace{-1mm}
\begin{equation*}
    \begin{aligned}
    \mathcal{L}_{max}(\ell(\mathcal{D}))=\max_{i\in \mathbb{N}_n}\ell_i=\ell_{[1]}.
    \end{aligned}
    \vspace{-1mm}
\end{equation*}
It is obvious that the maximum loss also belongs to the rank-based aggregate loss because it only considers the largest (top-1) individual loss during learning. 

Problems of maximum loss play significant roles in optimization and machine learning. When $\ell_i$ is a linear function, minimizing the maximum loss is equivalent to hard-margin SVM training (with $\ell_i$ representing the negative margin on the $i$-th sample) \cite{clarkson2012sublinear, hazan2011beating}. Beyond the linear case, minimizing the maximum loss can improve training speed and generalization, especially when dealing with rare informative samples, as argued in \cite{shalev2016minimizing}.  Furthermore, it is also the basic paradigm of robust optimization \cite{ben2002robust, namkoong2016stochastic}. Several works follow \cite{shalev2016minimizing} to discuss the optimization of maximum loss since the learning objective based on the maximum loss is known as the minimax learning \cite{lan2020first}. For example, the goal of \cite{chen2017robust} is to find a minimax solution that optimizes in the worst case over all individual losses. A recent work \cite{carmon2021thinking} addresses the complexity of minimizing the maximum loss for convex, Lipschitz, and non-smooth functions of $\ell_i$. In \cite{mohri2019agnostic}, the maximum loss is applied to subsets of the dataset instead of single data points in federated learning. \shu{Several typical loss functions are shown in Table \ref{table:maximum}.}

\smallskip
\noindent
\shu{\textbf{Discussion}. The maximum aggregate loss is beneficial in detecting and addressing extreme individual losses, which can speed up the model training process. However, it gives too much weight to the worst-performing sample. This may result in overfitting the outliers and not accurately representing the model's  performance on the entire dataset.}

\vspace{-1mm}

\begin{table}[h]
\centering
\scalebox{0.94}{
\begin{tabular}{l|l|l}
\hline
Work & Connection & Form \\ \hline
\cite{shalev2016minimizing} & Classification& $\max_{i\in \mathbb{N}_n}\mathbb{I}_{[y_if(\x_i)<1]}$ \\ \hline
\cite{chen2017robust}& Regression& $\max_{i\in \mathbb{N}_n}\|y_i-f(\x_i)\|_2^2$\\ \hline
\end{tabular}
}
\vspace{-0.5em}
\caption{\em \shu{Typical loss functions of the maximum aggregate loss.}
}
\label{table:maximum}
\vspace{-0.5cm}
\end{table}

\vspace{-3mm}

\subsection{Median Aggregate Loss}
\vspace{-1mm}
The average aggregate loss has limitations and is not robust against outliers. To address this issue, the median aggregate loss can be used for more robust mean estimation  \cite{shibzukhov2021principle, shibzukhov2022minimizing}. The median aggregate loss is calculated as follows:
\vspace{-1mm}
\begin{equation*}
    \begin{aligned}
    \mathcal{L}_{med}(\ell(\mathcal{D}))={1 \over 2}
    \left(\ell_{\left[\lfloor \frac{n+1}{2}\rfloor\right]} + \ell_{\left[\lceil \frac{n+1}{2} \rceil\right]}\right).
    \end{aligned}
    \vspace{-1mm}
\end{equation*}
We have $\mathcal{L}_{med}(\ell(\mathcal{D}))=\ell_{[\frac{n+1}{2}]}$ if $n$ is odd and $\mathcal{L}_{med}(\ell(\mathcal{D}))=\frac{1}{2}(\ell_{[\frac{n}{2}]}+\ell_{[\frac{n+2}{2}]})$ if $n$ is even.  \shu{Several typical loss functions are shown in Table \ref{table:median}.}

\textbf{Connect with least median squares (LMS)}. The median loss is also widely used in regression and classification due to its robustness against outliers. For example, 
In regression, the least median squares (LMS) \cite{rousseeuw1984least, rousseeuw1984robust}method is commonly used.
Similarly, in classification, a robust class conditional median loss is proposed in \cite{ma2011robust}, which uses the summation of the median individual losses in each class. However, optimizing the median loss directly is not possible due to its lack of differentiability. To overcome this, an iterative reweighting scheme is proposed in \cite{shibzukhov2021principle, shibzukhov2022minimizing}.

\textbf{Connect with median-of-means (MOM)}. The median-of-means (MOM) estimator, proposed by \cite{nemirovskij1983problem}, is a robust method to handle outliers. It can be regarded as a variant of median aggregate loss. The MOM estimator works by first randomly dividing N samples into M batches and then computing the mean for each batch. Finally, the median of these batch means is outputted. Recently, MOM estimators have been used in high-dimensional robust regression  \cite{brownlees2015empirical, hsu2016loss, jalal2020robust, lecue2020robust, lugosi2019risk} by applying MOM estimator on the loss function of empirical risk minimization process. A recent work \cite{liu2022robust} proposes a robust imitation learning model by minimizing a MOM-based learning objective.

\smallskip
\noindent
\shu{\textbf{Discussion}. The median aggregate loss is useful in scenarios where the data has outliers, but it may be less efficient and challenging to optimize.}

\vspace{-1mm}

\begin{table}[h]
\centering
\scalebox{0.84}{
\begin{tabular}{l|l|l}
\hline
Work & Connection & Form \\ \hline
\cite{shibzukhov2021principle} & LMS& \makecell[l]{$med\{\ell_1,...,\ell_n\}$, $\ell_i=(y_i-f(x_i))^2$,\\ \text{$med$: Median operation}} \\ \hline
\cite{hsu2016loss}  & MOM & \makecell[l]{\text{Randomly partition $\{\ell_1,...,\ell_n\}$ into $k$ subsets.}\\ \text{Let $u_i$ be the mean of $i$-th subset. }\\$med\{u_1,...,u_k\}$, $\ell_i=\frac{1}{2}(y_i-f(\x_i))^2$} \\ \hline
\end{tabular}
}
\vspace{-0.5em}
\caption{\em \shu{Typical loss functions of the median aggregate loss.}
}
\label{table:median}
\vspace{-0.5cm}
\end{table}

\vspace{-2mm}

\subsection{Average Top-$k$ (AT$_k$) Aggregate Loss}\label{sec:atk}
\vspace{-1mm}
The median loss can solve the issue of outliers, but it cannot address imbalanced data situations, and its learning objective is often non-convex. 
To mitigate these drawbacks, the average top-$k$ (AT$_k$) loss \cite{fan2017learning} is proposed, which is the average of the largest $k$ individual losses, that is defined as:
\vspace{-1mm}
\begin{equation*}
    \begin{aligned}
    \mathcal{L}_{at-k}(\ell(\mathcal{D}))=\frac{1}{k}\sum_{i=1}^k\ell_{[i]},
    \end{aligned}
    \vspace{-1mm}
\end{equation*}
where $1\leq k\leq n$. We can find that the AT$_k$ loss generalizes the average loss ($k=n$) and the maximum loss ($k=1$). Therefore, it can adapt to imbalanced and/or multi-modal data distributions better than the average loss and is less sensitive to outliers than the maximum loss. \shu{Several typical loss functions are shown in Table \ref{table:atk}.}

Since the AT$_k$ loss involved the sorting operation, it will bring a high time complexity in the training when directly optimizing it. Therefore, \cite{fan2017learning} proposes a reformulation of the AT$_k$ loss as the minimization on the average of the individual losses over all training examples transformed by a hinge function: 
$\mathcal{L}_{at-k}(\ell(\mathcal{D}))=\min_{\lambda}\lambda+\frac{1}{k}\sum_{i=1}^n[\ell_i-\lambda]_+$.

\textbf{Connect with condition value at risk (CVaR)}. Many works \cite{williamson2019fairness, curi2020adaptive, lee2020learning, laguel2021superquantiles, holland2021spectral, laguel2022superquantile} study and prove that the reformulated AT$_k$ loss is equivalent to the risk measure called CVaR \cite[Chapter~6]{ shapiro2021lectures} in portfolio optimization for effective risk management, which leads another reason why AT$_k$ loss is a robust learning approach for training machine learning models.
The work \cite{williamson2019fairness} also studies the connection between AT$_k$ loss and fairness based on CVaR. Robey et al. \cite{robey2022probabilistically} connect CVaR with Adversarial Training and then design a probabilistically robust learning model.  

\textbf{Connect with support vector machine (SVM)}. The AT$_k$ loss can be used with the hinge loss to generalize two convex SVMs, the C-SVM \cite{cortes1995support} and the $\nu$-SVM \cite{scholkopf2000new}. The extended $\nu$-SVM (E$\nu$-SVM) \cite{perez2003extension} is also related to AT$_k$ loss, as it adds only one nonconvex constraint to the $\nu$-SVM. Furthermore, \cite{takeda2008nu} shows that E$\nu$-SVM minimizes CVaR, which is exactly equivalent to minimizing  AT$_k$ loss.  

\textbf{Connect with distributionally robust optimization (DRO)}. Many studies, including \cite{lyu2020average, levy2020large, roux2021efficient}, link AT$_k$ loss with DRO \cite{rahimian2019distributionally}, which is a popular topic in operations research and statistical learning communities \cite{ben2002robust}. DRO can handle the data distribution shift problems between training and testing. \cite{zhu2019robust} uses the DRO version of AT$_k$ loss to create a robust game framework for pool-based active learning. \cite{qi2020simple} presents a DRO-TopK framework based on AT$_k$ loss for pairwise deep metric learning. \cite{sapkota2021distributionally} uses AT$_k$ as a constraint in a Gaussian processes mixture framework to develop a robust Deep Kernel Multiple Instance Learning model.

\textbf{Connect with learning strategies}. 
The SGD is a popular optimization method for deep learning models. However, all training samples may not equally important, and many can be ignored after a few epochs of training without affecting performance. To address this,  \cite{kawaguchi2020ordered} proposes ordered SGD based on AT$_k$ to learn deep learning models that improve testing performance. Inspired by ordered SGD, \cite{wu2020curricula} connects AT$_k$ loss to the large current loss minimization.
\cite{li2020tilted, li2021tilted} generalize the AT$_k$ loss with a tilted empirical risk minimization framework. In \cite{yuan2020group}, the authors employ the AT$_k$ loss as a classification error measure and propose AT$_k$ group sparse additive machine (AT$_k$-GSAM) for high-dimensional variable selection.

\textbf{Connect with deep learning applications}. The AT$_k$ loss is also used in various real-world scenarios. For example, \cite{xu2019mixed} combines the average loss and AT$_k$ loss for optic disc and optic cup segmentation with deep learning in imbalanced data. 
It is also applied to train deep learning models for sonar image generation \cite{lee2018deep,lee2019deep}, head pose estimation \cite{huang2020improving}, and 6D pose estimation based on rotational primitive reconstruction \cite{jeon2020prima6d}.
The authors in \cite{oh2020cnn} design a supervised learning approach based on AT$_k$ loss for identifying ChIP-seq (one of the core experimental resources available to understand genome-wide epigenetic
interactions and identify the functional elements associated with diseases) using convolutional neural networks (CNNs). A top-$k$ cross-entropy loss based on AT$_k$ loss is proposed in \cite{wang2021simultaneous} to design a  multi-task learning framework for simultaneously identifying the right ventricle end-diastolic and end-systolic frames
and detecting anatomical landmarks for echocardiography. The top-$k$ aggregate loss is also extended to generative adversarial network (GAN) models called top-$k$ GAN in \cite{sinha2020top}. It trains the generator by removing the elements of the mini-batch with the lowest critical outputs. By performing the top-$k$ operation on the predictions from the critic, the generator can generate samples close to the target distribution and make them more realistic.

\smallskip
\noindent
\shu{\textbf{Discussion}. The AT$_k$ aggregate loss can be useful in reducing the effect of outliers and addressing imbalanced datasets. However, it cannot completely eliminate the influence of the outliers.}

\vspace{-1mm}
\begin{table}[h]
\centering
\scalebox{0.84}{
\begin{tabular}{l|l|l}
\hline
Work & Connection & Form \\ \hline
\cite{fan2017learning} & CVaR& \makecell[l]{$\frac{1}{k}\sum_{i=1}^k\ell_{[i]}=\min_{\lambda\in\mathbb{R}}\frac{1}{k}\sum_{i=1}^n[\ell_i-\lambda]_++\lambda$,\\ \text{where} $\ell_i=[1-y_if(\x_i)]_+$} \\ \hline
\cite{scholkopf2000new}& SVM& \makecell[l]{$\min_{\lambda\in\mathbb{R}}\frac{1}{n}\sum_{i=1}^n[\ell_i-\lambda]_++\frac{k}{n}\lambda+\frac{1}{2}\|f\|^2$,\\ \text{where} $\ell_i=[1-y_if(\x_i)]_+$ and $\|\cdot\|^2$ is a norm}\\ \hline
\cite{levy2020large}  & DRO & \makecell[l]{$\frac{1}{k}\sum_{i=1}^k\ell_{[i]}=\sup_{q\in\Delta^n}\{\sum_{i=1}^n q_i\ell_i \ \text{s.t.} \|q\|_\infty\leq \frac{1}{k}\}$ \\ where $\Delta^n=\{q\in \mathbb{R}^n_{+}|q^\top \mathbf{1}=1\}$} \\ \hline
\cite{kawaguchi2020ordered} & SGD & \makecell[l]{Find a set $Q$ of top-$q$ individual losses\\ in a mini-batch of samples. Then use $\frac{1}{q}\sum_{i\in Q}\ell_{i}$.} \\ \hline
\end{tabular}
}
\vspace{-0.5em}
\caption{\em \shu{Typical loss functions of the AT$_k$ aggregate loss.}
}
\label{table:atk}
\vspace{-0.5cm}
\end{table}

\vspace{-2mm}

\subsection{Average Bottom-$k$ (AB$_k$) Aggregate Loss} 
\vspace{-1mm}
The AT$_k$ loss cannot completely eliminate the impact of outliers and noisy labels, which often have the highest individual losses. To address this issue, \cite{shen2019learning, shen2019iterative, shah2020choosing, roh2021sample} focus on the small losses during training. Based on these works, the average bottom-$k$ (AB$_k$) loss is introduced and defined as: 
\vspace{-1mm}
\begin{equation*}
    \begin{aligned}
    \mathcal{L}_{ab-k}(\ell(\mathcal{D}))=\frac{1}{k}\sum_{i=n-k+1}^n \ell_{[i]}.
    \end{aligned}
    \vspace{-1mm}
\end{equation*}
\shu{Several typical loss functions are shown in Table \ref{table:abk}.}

\textbf{Connect with the trimmed loss}. The least trimmed square (LTS) estimator is a robust estimator \cite{huber2004robust} for regression that minimizes the loss of only a fraction of samples, proposed by \cite{rousseeuw1984least}.  Building upon this idea, \cite{shen2019learning} propose the iterative LTS (ILTS) approach, which iteratively minimizes the $\mathcal{L}_{ab-k}$ loss to obtain an optimal classifier in the presence of corrupted training samples.  ILTS selects a subset of samples with the lowest individual loss and refits the model only on that subset.  ILTS is also connected to GAN models, where it can be used to train the critic on selected samples with small discriminator's loss, corresponding to learning with the bottom $k$ discriminator's losses when some fraction of training samples are from a lousy dataset.

Based on ILTS, the authors in \cite{shen2019iterative} apply ILTS to solve mixed linear regression with adversarial corruptions. The Min-$k$ SGD (MKL-SGD) proposed in \cite{shah2020choosing} has a similar idea to ILTS, which also belongs to the AB$_k$ learning strategy. In each iteration of MKL-SGD, the algorithm selects a set of $k$ samples and updates the model parameters using the samples with the smallest individual losses. In \cite{yuan2020learning}, the authors use ILTS to learn entangled single-sample distribution by estimating a common parameter shared by a family of distributions, given one single sample from each distribution.  To tackle noisy-labeled data in DNNs, the authors of \cite{song2020no} propose an adaptive $k$-set selection approach, which selects $k$ samples with a small noise risk from the whole noisy training samples at each training epoch. This approach can also be regarded as a type of AB$_k$ strategy. To understand why the DNNs using the small-loss criterion lean well from noisy labels, in \cite{gui2021towards}, the authors formalize the small-loss criterion to better tackle noisy labels and view this criterion from a theoretical perspective.

\textbf{Connect with fairness}. AB$_k$ is also used in recent research to develop fair and robust models, such as in \cite{roh2021sample}. In their work, the authors aim to learn a robust model from corrupted data while also addressing fairness. To achieve this, they add a fairness constraint called demographic parity to the AB$_k$ loss to formulate the learning objective.

\textbf{Connect with inverted CVaR}. In \cite{lee2020learning}, the authors connect AB$_k$ with CVaR by proposing an inverted version of CVaR.  They show that the AB$_k$ loss is equivalent to the inverted CVaR, which is used to describe algorithms that ignore high-loss examples.  Similarly, Han et al. \cite{han2018co} propose a learning framework to handle noisy labels by training two models simultaneously and selecting and feeding a fraction of samples with the lowest loss to each other.  Their objective function can be regarded as an inverted CVaR or AB$_k$ loss.

\textbf{Connect with unsupervised learning}. AB$_k$ can also be used in designing robust unsupervised learning models. As mentioned by Hampel \cite{hampel2001robust}, outliers do not fit the pattern set by the majority of the data. The authors in \cite{maurer2021robust} suggest focusing on a sufficient portion of the data during the model training. To this end, they propose applying L-estimators \cite{serfling2009approximation}, which are formed by a weighted average of the order statistics. Given a candidate learning model, they first rank its losses on the training data and then take a weighted average emphasizing small losses more. In other words, this construction can be the average of a fraction of the smallest losses, which is exactly the AB$_k$ loss. They use the weighted AB$_k$ loss to design a robust $k$-Means clustering model and study the robustness of Principal Subspace Analysis. 

\smallskip
\noindent
\shu{\textbf{Discussion}. The AB$_k$ aggregate loss can be useful in addressing outliers and improving the stability of the training process. However, it may slow down the convergence speed of model training and overlook the significant individual losses. This can make it difficult to obtain a precise model.}

\vspace{-1mm}
\begin{table}[h]
\centering
\scalebox{0.84}{
\begin{tabular}{l|l|l}
\hline
Work & Connection & Form \\ \hline
\cite{shen2019learning} & LTS& $\sum_{i=n-k+1}^n \ell_{[i]}$, where $\ell_i=(y_i-f(\x_i))^2$ \\ \hline
\cite{roh2021sample}& Fairness& \makecell[l]{$\sum_{i=1}^np_i \ell_i$ \\ s.t. $\sum_{i=1}^n p_i\leq k$, $p_i\in\{0,1\},$ fairness constraints}\\ \hline
\cite{lee2020learning}  & Inverted CVaR & \makecell[l]{$\frac{1}{k}\sum_{i=n-k+1}^n \ell_{[i]}=\max_{\lambda\in\mathbb{R} }\lambda-\frac{1}{k}\sum_{i=1}^n[\lambda-\ell_i]_+$} \\ \hline
\cite{kawaguchi2020ordered} & L-estimators & \makecell[l]{$\frac{1}{n}\sum_{i=1}^n W(\frac{i}{n})\ell_{[i]}$, where $W$ is a non-increasing \\ and zero on $[\xi,1]$ for $\xi<1$ }  \\ \hline
\end{tabular}
}
\vspace{-0.5em}
\caption{\em \shu{Typical loss functions of the AB$_k$ aggregate loss.}
}
\label{table:abk}
\vspace{-0.5cm}
\end{table}

\vspace{-3mm}

\subsection{Average of Ranked Range (AoRR) Aggregate Loss}
\vspace{-1mm}
As we mentioned, the average loss is insensitive to minority sub-groups, while the maximum loss is sensitive to outliers.
The AT$_K$ can dilute but not exclude the influences of the outliers. The AB$_k$ is also insensitive to minority sub-group data since they focus on more samples with lower loss values. To this end, the average of ranked range (AoRR) loss is proposed in \cite{hu2020learning}. Unlike previous aggregate losses, the AoRR loss is robust to imbalanced data and can completely eliminate the influence of outliers if their proportion in training data is known. It can be defined as:
\vspace{-1mm}
\begin{equation*}
    \begin{aligned}
    \mathcal{L}_{aorr}(\ell(\mathcal{D}))
    = \frac{1}{k-m}\sum_{i=m+1}^k \ell_{[i]},
    \end{aligned}
    \vspace{-1mm}
\end{equation*}
where $0\leq m<k\leq n$. The AoRR loss can be generalized to the average loss ($k=n$ and $m=0$), the maximum loss ($k=1$ and $m=0$), the median loss ($k=\lceil \frac{n+1}{2} \rceil$ and $m= \lfloor \frac{n+1}{2}\rfloor-1$), the AT$_k$ loss ($m=0$), and the AB$_k$ loss ($k=n$, $m=n-k$). In addition, the robust version of the maximum loss  \cite{shalev2016minimizing}, which is a maximum loss on a subset of samples of size at least $n-(k-1)$, where the number of outliers is at most $k-1$, is equivalent to the top-$k$ loss, a special case of the AoRR aggregate loss ($m=k-1$). \shu{Several typical loss functions are shown in Table \ref{table:aorr}.}

\textbf{Connect with bilevel optimization}. In \cite{hu2021sum}, the authors show that the AoRR loss can be formulated as a cardinality-constrained bilevel optimization problem. This connection allows them to adapt existing bilevel optimization algorithms from \cite{borsos2020coresets} to solve the problem.

\textbf{Connect with interval CVaR}. Hu et al. \cite{hu2021sum} also reformulate AoRR as the difference between two sums of the top-ranked values such that $\mathcal{L}_{aorr}(\ell(\mathcal{D}))
= \frac{1}{k-m}[\sum_{i=1}^k \ell_{[i]} - \sum_{i=1}^m \ell_{[i]}]$. They represented each sum of the top-ranked value as a variant of AT$_k$ loss, which is equivalent to the formula of CVaR discussed in Section \ref{sec:atk}. Thus, they further reformulate AoRR loss to the difference between two CVaRs, which is named Interval Condition Value at Risks (ICVaRs). A recent work \cite{liu2022risk} further statistical studies ICVaR for robust regression and classification in nonsmooth settings. 

\textbf{Connect with the difference of convex optimization}. As we mentioned, the AoRR loss is a difference of top sums of top-ranked values, and each of them is a convex optimization problem when using a convex individual loss function. This formulation leads to a difference of convex (DC) programming problem \cite{le2018dc}, discussed in \cite{hu2020learning}. Therefore it can be effectively solved via DC algorithm (DCA) \cite{tao1986algorithms}. In addition, Yao et al. \cite{yao2022large} apply the AoRR loss and DCA to optimize partial AUC \cite{narasimhan2013svmpauctight}.  

\textbf{Connect with SVM}. When the $\ell$ individual loss function is a hinge loss function that works on a linear model and constrains the $l_2$ norm of the model parameters to 1, the AoRR loss becomes the extended robust SVM (ER-SVM), which is proposed in \cite{fujiwara2017dc}. The authors also proved that the ramp-loss SVM \cite{collobert2006trading} and the robust outlier detection (ROD) method \cite{xu2006robust} are special cases of ER-SVM, which means AoRR can also generalize to these two methods. A similar model, CVaR-($\alpha_L$, $\alpha_U$)-SVM, which relaxes the constraint of the $l_2$ norm of the model parameters less than 1, was proposed in \cite{tsyurmasto2013support}. Note that $\alpha_L$, $\alpha_U$ are the same as $k$, $m$, which control the range size. Without constraint, the ER-SVM will be reduced to ($\nu$, $\mu$)-SVM \cite{kanamori2017breakdown, kanamori2017robustness}, which is exactly equivalent to AoRR using hinge loss function and has been theoretically evaluated the robustness for outliers. Similarly, $\nu$ and $\mu$ have the same properties of $k$ and $m$. 

\textbf{Connect with the trimmed root mean squared
error (tRMSE)}. When $\ell$ is a mean squared error loss, the AoRR loss becomes the tRMSE, which is used to measure the performance of the recommender system in \cite{ortis2019predicting}. 

\smallskip
\noindent
\shu{\textbf{Discussion}. The AoRR aggregate loss can generalize to many existing aggregate losses, making use of their advantages. It is also robust to outliers and imbalanced data. However, it is a nonconvex loss, which can make it challenging to optimize and obtain the optimal model parameters.}

\vspace{-1mm}

\begin{table}[h]
\centering
\scalebox{0.80}{
\begin{tabular}{l|l|l}
\hline
Work & Connection & Form \\ \hline
\cite{hu2021sum} & \makecell[l]{Bilevel \\optimization}& \makecell[l]{$\min_{\lambda,q}(k-m)\lambda+\sum_{i=1}^nq_i[\ell(f^*(\x_i),y_i)-\lambda]_+$\\ s.t. $f^*\!\in\! \arg\!\min_f \!(k\!-\!m)\lambda\!+\!\sum_{i=1}^nq_i[\ell(f(\x_i),y_i)-\lambda]_+$\\ $q_i\in[0,1], \|q\|_0=n-m$} \\ \hline
\cite{hu2021sum}& \makecell[l]{Interval CVaR}& \makecell[l]{$\frac{1}{k-m}[\sum_{i=1}^k \ell_{[i]} - \sum_{i=1}^m \ell_{[i]}]$ }\\ \hline
\cite{hu2020learning}  & \makecell[l]{DC \\optimization} & \makecell[l]{$\frac{1}{k-m}\{\min_{\lambda}k\lambda+\sum_{i=1}^n[\ell_i-\lambda]_+-$\\$\min_{\lambda'}m\lambda'+\sum_{i=1}^n[\ell_i-\lambda']_+\}$} \\ \hline
\cite{kawaguchi2020ordered} & SVM & \makecell[l]{$\min_{\rho\in\mathbb{R}, \eta\in E_m} \nu\rho+\frac{1}{n}\sum_{i=1}^n\eta_i[\ell_i-\rho]_+$, where $0\!<\!\nu\!<\!1$\\ $E_m=\{(\eta_1,...,\eta_n)\in\{0,1\}^n:\sum_{i=1}^n \eta_i=n-m\}$ }  \\ \hline
\end{tabular}
}
\vspace{-0.5em}
\caption{\em \shu{Typical loss functions of the AoRR aggregate loss.}
}
\label{table:aorr}
\vspace{-0.5cm}
\end{table}

\vspace{-2mm}

\subsection{ Close-$k$ Aggregate Loss}\label{sec:close-k}
\vspace{-1mm}
When the training data contains ambiguous samples that cannot be classified correctly, the AT$_k$ aggregate loss cannot make an optimal classification. The reason is that the AT$_k$ loss is still rewarded for reducing the loss of the ambiguous samples, even if these samples cannot be classified correctly. To address this issue, He et al. in \cite{he2018minimizing} propose a close-$k$ aggregate loss, which is defined as:
\vspace{-1mm}
\begin{equation*}
    \begin{aligned}
    &\mathcal{L}_{close-k}(\ell(\mathcal{D}))\\
    =&\sum_{i=1}^n \begin{cases}
\ell_{[n-i+1]} & \text{ if } i\leq k \\
0 & \text{ if } i>k \ \text{and} \ \ell_{[n-i+1]} \text{correctly  classified} \\
M & \text{ if } i>k \ \text{and} \ \ell_{[n-i+1]} \text{incorrectly  classified} 
\end{cases},
    \end{aligned}
    \vspace{-1mm}
\end{equation*}
where $M$ is a constant and $\ell_{[n-i+1]}$ is the individual loss $\ell_j$ with the ($n-i+1$)-th largest $|\ell_j-T|$ such that T is another constant. By applying this operation, they can select the individual loss of the ($n-i+1$)-th closest sample to the decision boundary.

\smallskip
\noindent
\shu{\textbf{Discussion}. The Close-$k$ aggregate loss can handle ambiguous samples in the training set. However, it requires tuning of three hyperparameters, which can be challenging. Furthermore, its application is limited to specific scenarios.}

\vspace{-3mm}

\section{Rank-based Individual Losses}\label{sec:individual_loss}
\vspace{-1mm}
In this section, we will introduce four types of rank-based individual losses and 
show their broad application in machine learning.
\vspace{-3mm}


\subsection{Average Individual Losses} 
\vspace{-1mm}
Similar to the aggregate loss, the average operator is the traditional aggregator that appears in the individual loss. We can represent the average individual loss as:
\vspace{-1mm}
\begin{equation*}
    \begin{aligned}
    \mathcal{L}_{avg}(\mathbb{O}(f(\x)))= \frac{1}{C}\sum_{j\in\mathbb{N}_C}\mathcal{O}(f_j(\x),\y)=\frac{1}{C}\sum_{j=1}^C\mathcal{O}(f_{[j]}(\x),\y),
    \end{aligned}
    \vspace{-1mm}
\end{equation*}
where $\mathcal{O}(f_{[j]}(\x),\y)$ represents the top-$j$ value in  $\mathbb{O}(f(\x))$. \shu{Several typical loss functions are shown in Table \ref{table:average-individual}.}

\textbf{Connect with multi-class classification}. 
In \cite{weston1999support}, a novel approach for multi-class SVM classification was proposed, which formulates the problem as a single quadratic program. This approach considers all class relationships simultaneously, using a sum operator applied to relative margins in $\mathbb{O}(f(\x))$, and can be expressed as $C\cdot\mathcal{L}_{avg}(\mathbb{O}(f(\x)))$. 
Independently, two similar methods were proposed around the same time in \cite{bredensteiner1999multicategory} and \cite[Section 10.10]{vapnik1999nature}, inspired by the work in \cite{weston1999support}. 
To build a multi-category classifier using binary classifiers, one popular method is called one-versus-all (OVA). OVA uses $C$ different binary decision functions, each separating one class from all the rest \cite[Section 10.10]{vapnik1999nature}. Thus, OVA sums all binary problems into one learning problem, which can be seen as a sum aggregation that adapts the average aggregator.

Lee et al. \cite{lee2004multicategory} proposed an alternative approach for multi-class SVM called the LLW method. This method uses absolute margins in an all-in-one approach and combines the absolute margin violations using the average aggregator. Liu et al. \cite{liu2011reinforced} introduced another method, called Reinforced Multi-Category SVM (RM-SVM), which also uses the average aggregator to combine margins. However, in RM-SVM, the margins are weighted based on the ground truth.

\textbf{Connect with multi-label classification}. The average aggregator is also widely used in multi-label classification. For example, the instance-F1 loss \cite{zhang2013review} applies the average aggregator to calculate F-measure \cite{sasaki2007truth} based on all labels, and it can be decomposable.
The Hamming loss \cite{schapire2000boostexter, zhang2013review} reduces multi-label learning to $C$ independent binary classification problems (one for each label) and ignores label dependencies that are helpful to multi-label learning. It is a typical example that uses the average aggregator.

\smallskip
\noindent
\shu{\textbf{Discussion}. The average individual loss considers all label prediction scores, but it may overlook important patterns or trends between labels that could be utilized to enhance the model's performance.}

\vspace{-1mm}
\begin{table}[h]
\centering
\scalebox{0.80}{
\begin{tabular}{l|l|l}
\hline
Work & Connection & Form \\ \hline
\cite{weston1999support} & \makecell[l]{Multi-class}& \makecell[l]{$\sum_{j\in \mathbb{N}_C:j\neq \y}[1+f_j(\x)-f_{\y}(\x)]_+$} \\ \hline
\cite{zhang2013review}& \makecell[l]{Multi-label}& \makecell[l]{$\frac{1}{C}\sum_{j\in \mathbb{N}_C}[\mathbb{I}_{j\notin \y}\cdot \mathbb{I}_{f_j(\x)\geq 0.5}+\mathbb{I}_{j\in \y}\cdot \mathbb{I}_{f_j(\x)< 0.5}]$}\\ \hline
\end{tabular}
}
\vspace{-0.5em}
\caption{\em \shu{Typical loss functions of the average individual loss.}
}
\label{table:average-individual}
\vspace{-0.5cm}
\end{table}
\vspace{-2mm}

\subsection{Maximum Individual Loss}
\vspace{-1mm}
Many multi-class and multi-label classification problems design individual losses based on the maximum aggregator. Following the notations that we defined before, we can represent the maximum guided individual loss as:
\vspace{-1mm}
\begin{equation*}
    \begin{aligned}
    \mathcal{L}_{max}(\mathbb{O}(f(\x)))= \max_{j\in\mathbb{N}_C}\mathcal{O}(f_j(\x),\y)=\mathcal{O}(f_{[1]}(\x),\y),
    \end{aligned}
    \vspace{-1mm}
\end{equation*}
where $\mathcal{O}(f_{[1]}(\x),\y)$ represents the top-$1$ value in  $\mathbb{O}(f(\x))$. \shu{Several typical loss functions are shown in Table \ref{table:max-individual}.}

\textbf{Connect with multi-class classification}. In binary classification, the 0-1 loss is the standard performance measure. However, because it is difficult to optimize, surrogate losses such as the SVM hinge loss are often used instead. Researchers have extended these surrogate losses to solve multi-class classification problems as the number of classes in the data set increases. One popular surrogate loss is the multi-class SVM, proposed by Grammer and Singer \cite{crammer2001algorithmic} and  $\mathcal{O}(f_j(\x),\y)$ can be defined as $[1+f_j(\x)\times \mathbb{I}_{j\neq \y}-f_{\y}(\x)]_+$ or $[\mathbb{I}_{j\neq\y}+f_j(\x)-f_{\y}(\x)]$. The loss becomes large if the ground truth label $\y$ score is below the largest score from other class labels. Dogan et al. \cite{dogan2016unified} proposed two similar losses, AMO and ATM, which are based on the maximum aggregator.

\textbf{Connect with multi-label classification}. The maximum aggregator can also be used for multi-label learning tasks where a sample can have multiple labels. One of the conventional multi-label losses  is proposed in \cite{crammer2003family} such that $\mathcal{O}(f_j(\x),\y)$ is represented as $[1+f_j(\x)\times \mathbb{I}_{j\notin \y}-\min_{y\in \y}f_y(\x)]_+$. This loss function requires the minimal prediction score from ground truth labels to be larger than the maximal score from non-ground truth labels. The multi-label loss can be reduced to the multi-class loss if $|\y|=1$. This loss is also known as the multiclass multilabel perceptron (MMP) \cite{furnkranz2008multilabel} and the separation ranking loss \cite{guo2011adaptive}.  One-error and Coverage losses \cite{schapire2000boostexter, zhang2013review} are also maximum-guided individual losses, where they penalize if the label of the top predictions does not belong to ground truth labels. 

\textbf{Connect with Adversarial Attacks}. Nowadays, DNNs have significantly improved, or in some cases, revolutionized, the state-of-the-art performance of many computer vision problems. Notwithstanding this tremendous success, the omnipotent DNN models are surprisingly vulnerable to adversarial attacks \cite{szegedy2013intriguing,goodfellow2014explaining,liu2016delving}. In particular, inputs with specially designed perturbations, commonly known as {\it adversarial examples}, can easily mislead a DNN model to make erroneous predictions. This motivates the explorations of algorithms generating adversarial examples \cite{carlini2017towards,moosavi2016deepfool,madry2017towards} as a means to analyze the vulnerabilities of DNN models and improve their security. 

Most existing adversarial attacking methods target multi-class classification problems. These methods often target the top prediction and aim to change it with perturbations. Therefore, the maximum guided individual loss is broadly designed and used. For untargeted attacks, DeepFool \cite{moosavi2016deepfool} is a generalization of the minimum attack under general decision boundaries by swapping labels from the top-$2$ prediction.  Following DeepFool, the work of \cite{moosavi2017universal} (UAP) aims to find universal adversarial perturbations independent of individual input images.  Both DeepFool and UAPs aim to make the top-1 predicted label from the perturbed sample different from the top-1 predicted label from the original sample. For targeted attacks, the CW method \cite{carlini2017towards} aims to enforce the  top-1 predicted label from the perturbed sample that is equivalent to the pre-defined label (which is different from the ground truth label) when designing the loss function.  

The authors of \cite{song2018multi} describe an adversarial targeted attack to multi-label classification, extending existing attacks to multi-class classification.  This method is further studied in \cite{zhou2020generating}, which transfers the problem of generating an attack to a linear programming problem.  To make the predictions of adversarial examples lying inside of the training data distribution, \cite{melacci2020can} propose a multi-label attack procedure with an additional domain knowledge-constrained classifier. These are all for multi-label learning. They use the maximum operator to design the individual loss that can make the targeted labels ranked before all ground truth labels.

\smallskip
\noindent
\shu{\textbf{Discussion}. The maximum individual loss can prioritize the most critical predicted score. However, it may neglect the other top predictions, which could be used to improve model robustness and vulnerability checking.}

\vspace{-1mm}
\begin{table}[h]
\centering
\scalebox{0.80}{
\begin{tabular}{l|l|l}
\hline
Work & Connection & Form \\ \hline
\cite{crammer2001algorithmic} & \makecell[l]{Multi-class}& \makecell[l]{$\max_{j\in\mathbb{N}_C}[1+f_j(\x)\times \mathbb{I}_{j\neq \y}-f_{\y}(\x)]_+$} \\ \hline
\cite{crammer2003family}& \makecell[l]{Multi-label}& \makecell[l]{$\max_{j\in\mathbb{N}_C}[1+f_j(\x)\times \mathbb{I}_{j\notin \y}-\min_{y\in \y}f_y(\x)]_+$}\\ \hline
\cite{carlini2017towards}& \makecell[l]{Adversarial \\Attack}& \makecell[l]{$\|\delta\|_2+c\cdot \max_{j\neq t}[f_j(\x+\delta)-f_{t}(\x+\delta)]_+$, where $t\in\mathbb{N}_C$ is \\ the targeted label and $t\neq \y$. $\delta$ is a  perturbation and $c>0$.}\\ \hline
\end{tabular}
}
\vspace{-0.5em}
\caption{\em \shu{Typical loss functions of the maximum individual loss.}
}
\label{table:max-individual}
\vspace{-0.5cm}
\end{table}

\vspace{-2mm}

\subsection{Top-$k$ Individual Loss} 
\vspace{-1mm}
Class overlap, multi-label nature of samples, and class ambiguity problems appear when the number of classes increases in image classification. Top-$k$ error is explored and studied when a predictor allows $k$ guesses instead of one and is not penalized for $k-1$ mistakes. It is regarded as a robust evaluation measure in the current research and competition, such as the top-1 to top-5 performances are evaluated in ImageNet challenge \cite{russakovsky2015imagenet}. In such a case, top-$k$ accuracy is an important metric that estimates whether the candidates include correct targets, which limits total performance. Therefore, the top-$k$ guided individual loss is proposed and can be defined as:
\vspace{-1mm}
\begin{equation*}
    \begin{aligned}
    \mathcal{L}_{top-k}(\mathbb{O}(f(\x)))= \mathcal{O}(f_{[k]}(\x),\y),
    \end{aligned}
    \vspace{-1mm}
\end{equation*}
where $\mathcal{O}(f_{[k]}(\x),\y)$ represents the top-$k$ value in  $\mathbb{O}(f(\x))$. \shu{Several typical loss functions are shown in Table \ref{table:Tk-individual}.}

\textbf{Connect with multi-class classification}. The top-$k$ SVM method is first proposed by Lapin et al. in \cite{lapin2015top} for the linear model. The main motivation for top-$k$ loss is to relax the penalty for making an error in the top-$k$ predictions for multi-class classification. In this case, the function $\mathcal{O}$ is just the traditional multi-class function that we have defined before. Inspired by this method, Yan et al. \cite{yan2018top} integrate multiple feature fusion into the top-$k$ multi-class framework to improve the classification performance. To mitigate the instance ambiguity problem in the image-sentence matching scenario, Zhang et al. \cite{zhang2019deep} extend top-$k$ SVM and propose a deep multi-modal network with a top-$k$ ranking loss for large-scale image-sentence matching with indistinguishable data. Moreover, the classes in hierarchical classification are formed as a structured hierarchy. So, the misclassification costs depend on the relation
between the correct class and the incorrect class in the
hierarchy. The combination of top-$k$ classification and hierarchical classification is first proposed in \cite{oh2017top}. In \cite{oh2017top}, the authors define a top-$k$ hierarchical loss function based on top-$k$ SVM loss and then provide the Bayes-optimal solution that minimizes the expected
top-$k$ hierarchical misclassification cost. Chzhen et al. \cite{chzhen2021set} view top-$k$ SVM from the set-value classification aspect. They classify it as a point-wise size control problem. Tan et al. in \cite{tan2019exact} formulate the top-$k$ multi-class classification problem as an $l_0$ norm minimization problem and consider an exact penalty method to solve it.

The practical success of the top-$k$ SVM method motivates the theoretical analysis of consistent top-$k$ classification, which is a property of top-$k$ error and its surrogate loss. The consistency of surrogate loss states that the learned classifier converges to the population optimal prediction in the infinite sample limit. However, as Yang et al. studied in \cite{yang2020consistency}, the top-$k$ SVM method does not satisfy this top-$k$ consistency property. Therefore, they proposed another top-($k+1$) loss, which satisfies the top-$k$ calibration (a necessary condition of top-$k$ consistency) and further satisfies top-$k$ consistency under mild conditions. It can be represented as $\mathcal{O}(f_{[k+1]}(\x),\y)$, where $\mathcal{O}(f_j(\x),\y):=[1+f_j(\x)-f_{\y}(\x)]_+$. However, the top-$k$ calibrated loss proposed by \cite{yang2020consistency} is not a smooth loss function. Therefore, it cannot be directly used for training a deep neural network. In addition, it cannot handle imbalanced data. So it is not robust for the long-tailed dataset. To this end, following \cite{yang2020consistency}, Garcin et al. \cite{garcin2022stochastic} propose a smooth version of $\mathcal{O}(f_{[k+1]}(\x),\y)$ and a new loss for imbalanced top-$k$ classification. Specifically, they proposed to smooth the top-$k$ calibrated loss $\mathcal{O}(f_{[k+1]}(\x),\y)$ with the perturbed optimizers method developed by \cite{berthet2020learning}. In other words, they add Gaussian noise into the logits predicted by neural networks. To handle imbalanced data, they follow \cite{cao2019learning} and replace the margin parameter 1 from the formulation $\mathcal{O}(f_j(\x),\y)$ with another margin parameter which is determined by the number of samples in training set with ground truth class.

\textbf{Connect with multi-label classification}. Inspired by \cite{yang2020consistency}, Hu et al. \cite{hu2020learning} extend the top-$k$ calibrated multi-class loss to multi-label learning. The motivation is that the classifier is expected to include as many true labels as possible in the top $k$ outputs during the training. They design a top-$k$ multi-label (TKML) loss, which can also be represented as $\mathcal{O}(f_{[k+1]}(\x),\y)$ but $\mathcal{O}(f_j(\x),\y):=[1+f_j(\x)-\min_{y\in\y}f_y(\x)]_+$, where $|\y|\geq 2$. It is clear that the TKML loss can generalize the conventional multi-class loss ($|\y|=k=1$) and the top-$k$ calibrated $k$-guesses multi-class classification \cite{yang2020consistency} ($1=|\y|\leq k < C$). \cite{li2017improving} leverages the top-$k$ in label decision to improve the pairwise ranking for multi-label image classification. In \cite{hu2021sum}, the authors proposed a TKML-AoRR loss, which combines the AoRR and the TKML methods. It can improve the robustness of top-$k$ multi-label learning in the face of outliers in samples and labels alike.

\textbf{Connect with Adversarial Attacks}. Realizing that only attacking the top predictions may not be effective, several works introduce attacks to the top-$k$ (for $k > 1$) predictions in a multi-class classification system. 
$k$Fool \cite{tursynbek2022geometry} and CW$^k$ \cite{zhang2020learning} extend the original DeepFool \cite{moosavi2016deepfool} and CW \cite{carlini2017towards} methods to exclude the truth label out of the top-$k$ predictions. Specifically, $k$Fool is based on a geometry view of the decision boundary between $k$ labels and the truth label in the multi-class problem. The UAP method is extended in \cite{tursynbek2022geometry} to top-$k$ Universal Adversarial Perturbations ($k$UAPs). In addition, the CW method is extended to a top-$k$ version known as CW$^k$ in \cite{zhang2020learning}. For multi-label learning, Hu et al., \cite{hu2021tkml} propose an untargeted adversarial attack (TKML-AP-U) loss for TKML learning. TKML-AP-U attack aims to only replace the top-$k$ labels with a set of arbitrary $k$ labels that are not true classes of the untampered input. Furthermore, they propose TKML-AP-Uv loss, which extends TKML-AP-U to a universal adversarial attack independent of input \cite{moosavi2017universal} so it can be shared by all instances. 

\smallskip
\noindent
\shu{\textbf{Discussion}. The Top-$k$ individual loss can address class ambiguity issues. However, since it is a nonconvex function, optimizing it can be a challenging task.}

\vspace{-1mm}
\begin{table}[h]
\centering
\scalebox{0.80}{
\begin{tabular}{l|l|l}
\hline
Work & Connection & Form \\ \hline
\cite{yang2020consistency} & \makecell[l]{Multi-class}& \makecell[l]{$[1+f_{[k+1]}(\x)-f_{\y}(\x)]_+$} \\ \hline
\cite{hu2020learning}& \makecell[l]{Multi-label}& \makecell[l]{$[1+f_{[k+1]}(\x)-\min_{y\in\y}f_y(\x)]_+$}\\ \hline
\cite{hu2021tkml}& \makecell[l]{Adversarial \\Attack}& \makecell[l]{$\|\delta\|_2^2+[\max_{j\in\y}f_j(\x+\delta)-f_{[k+1]}(\x+\delta)]$, \\ where $\delta$ is a perturbation.}\\ \hline
\end{tabular}
}
\vspace{-0.5em}
\caption{\em \shu{Typical loss functions of the Top-$k$ individual loss.}
}
\label{table:Tk-individual}
\vspace{-0.5cm}
\end{table}
\vspace{-2mm}

\subsection{Average Top-$k$ (AT$_k$) Individual Loss}
\vspace{-1mm}
One potential reason to introduce the average top-$k$ aggregator for individual loss is that the top-$k$ aggregator is nonconvex. Therefore, finding a globally optimal solution is computationally intractable. As we discussed in Section \ref{sec:prop_agg}, the average top-$k$ is a convex upper bound of top-$k$. So the average top-$k$ aggregator is widely used to replace the top-$k$ aggregator in learning tasks. The average top-$k$ guided individual loss can be defined as:
\vspace{-1mm}
\begin{equation*}
    \begin{aligned}
    \mathcal{L}_{at-k}(\mathbb{O}(f(\x)))=\frac{1}{k}\sum_{j=1}^k \mathcal{O}(f_{[j]}(\x),\y).
    \end{aligned}
    \vspace{-1mm}
\end{equation*}
\shu{Several typical loss functions are shown in Table \ref{table:ATk-individual}.}

\textbf{Connect with multi-class classification}. According to the above motivation, Lapin et al. \cite{lapin2015top} propose two top-$k$ SVM losses named top-$k$ SVM$^\alpha$ and top-$k$ SVM$^\beta$ to relax the original top-$k$ SVM method for optimization inspired by ranking losses in \cite{usunier2009ranking}. Both of them adopt the AT$_k$ aggregator to replace the Top-$k$ aggregator in the original top-$k$ SVM. In addition, they apply a stochastic dual-coordinate ascent algorithm to optimize the proposed losses. However, this algorithm relies on the scoring method, which yields a high time complexity. Therefore, the following work \cite{chu2018optimizing} leverages the semismoothness of the problem and proposes to use the semismooth Newton algorithm for solving top-$k$ SVM$^\alpha$ and improve the training speed. 
Due to the above two losses being non-smooth, in their following works \cite{lapin2016loss, lapin2017analysis}, they propose different smooth versions of both losses based on Moreau-Yosida regularization \cite{nesterov2005smooth}. They conduct the experiments on linear models or pre-trained deep networks that are fine-tuned. However, the proposed losses cannot be directly used for training deep neural networks from a random initialization because of the non-smoothness of the top-$k$ SVM loss part and the sparsity of its gradient. Therefore, the following work \cite{berrada2018smooth} introduces another smooth version of top-$k$ SVM$^\alpha$ with a temperature parameter for deep top-$k$ classification. The main principle of the proposed loss is based on rewriting the top-$k$ SVM loss \cite{lapin2015top} as a difference of two max with logsumexp and using a divide-and-conquer approach to make the loss tractable. They mentioned the importance of smoothness and gradient density of loss in applying it to DL for successful optimization.
The work of \cite{rawat2020doubly} proposes a doubly-stochastic mining method, which combines the methods of ordered SGD from \cite{kawaguchi2020ordered} and top-$k$ SVM$^\beta$ from \cite{lapin2015top}. They use the average top-$k$ methods both in the data sample level and label level to construct the final loss function. Then the constructed loss is applied to solve modern retrieval problems, which are characterized by training sets with a vast number of labels and heterogeneous data distributions across sub-populations.

Another work in \cite{chang2017robust} proposes a robust top-$k$ SVM based on the top-$k$ SVM$^\alpha$ to address the outliers by using a hyperparameter to cap the values of the individual losses for visual category recognition.
Gong et al. \cite{gong2021top} extend top-$k$ SVM$^\alpha$ to deal with ambiguities in partial label learning (PLL) that aims to train a classifier from partially labeled data in order to automatically predict the ground truth label for an unseen instance. The proposed top-$k$ partial loss and convex top-$k$ partial hinge loss work well on partial label classification. 
Zhang et al. \cite{zhang2019deep} apply the top-$k$ SVM$^\alpha$ method to solve the image-sentence matching problem. Zhu et al. \cite{zhu2021unified} propose another top-$k$ calibrated loss named label-distributionally robust (LDR)-$k$ loss, which converts the top-$k$ SVM$^\beta$ method to a distributionally robust optimization problem and introduces a strongly convex regularizer. 
The above losses can be regarded as the unweighted average top-$k$ SVM. In addition, two weighted top-$k$ SVMs based on top-$k$ SVM$^\alpha$ and top-$k$ SVM$^\beta$ are proposed in \cite{kato2019learning, tajima2021frank}.  However, all of these losses cannot scale to extremely large output spaces because computing the value of them are required computing scores of all labels. To extend these losses to a large output space, the ordered weighted losses (OWLs) are proposed by Reddi et al. in \cite{reddi2019stochastic}. They also provide a stochastic negative mining approach to optimize the OWLs, which randomly sample a few classes other than the given positive class and treat them as negative classes with some additional correction while computing the loss based on the negative sampling approach from \cite{mikolov2013distributed}. 

Meantime, Lapin et al. \cite{lapin2016loss, lapin2017analysis} find the softmax loss only aims at top-1 performance and produces a reasonably good solution in top-1 error but ignores the top-2 error. Thus, they adapt the softmax loss to top-$k$ error optimization for multi-class classification. Then they propose two new average top-$k$ losses named top-$k$ entropy loss and truncated top-$k$ entropy loss for multi-class classification based on the softmax loss. They show that cross-entropy (CE) loss is top-$k$ calibrated for any $k$, which means the minimizers of CE will minimize the top-$k$ error for all $k$ under ideal conditions. This may be why CE loss succeeds in the top-$k$ performance of models in deep learning with large datasets.  The proposed losses are applied to the application of App usage prediction in \cite{zhao2019appusage2vec}. In particular, the authors in \cite{zhao2019appusage2vec} adjust the entropy loss by introducing hinge loss on top $k$. Such losses are also applied to solve egocentric action anticipation problems in \cite{furnari2018leveraging}. Later, Sawada et al. \cite{sawada2020trade} analyze that CE loss penalizes top-$k$ correct predictions similarly to incorrect predictions when it is incorrect in top-1 predictions. This leads to a sacrifice in top-$k$ correct prediction on samples challenging to predict correctly in top-1 prediction, instead of achieving top-1 correct prediction on other samples, to minimize the average of losses over all training samples. Therefore, CE is potentially not the best choice to obtain the best top-$k$ accuracy of all
possible predictions within practical limitations. Thus, they propose top-$k$ grouping loss, which contains two parts. The first part groups top-$k$ classes as a single class and apply CE loss to it. The second part is the traditional CE loss on the remaining classes. To keep a stable training performance, they define a top-$k$ transition loss that uses CE loss at the beginning of training and gradually transitions from CE loss to top-$k$ grouping loss.

\textbf{Connect with multi-label classification}. Groupwise ranking loss (GRLS) \cite{fan2020groupwise} is proposed for multi-label learning that implements the top-$k$ label principle. The GRLS loss is defined as the difference between the average predicted relevancy score of predicted positive labels (\ie, labels with top-$k$ largest prediction scores) and the average predicted relevancy score of ground truth positive labels. Therefore, we can set $k=|\y|$ and $\mathcal{O}(f_{j}(\x),\y) := f_{j}(\x)-\frac{1}{|\y|}\sum_{y\in\y}f_y(\x)$ to construct the GRLS loss based on $\mathcal{L}_{at-k}(\mathbb{O}(f(\x)))$. This loss is straightforward to understand. It tries to push the ground truth labels to the top list according to the value of the prediction score. Following the smooth top-$k$ entropy loss and truncated top-$k$ entropy loss from \cite{lapin2016loss}, Amos et al. \cite{amos2019limited} propose the limited multi-label (LML) projection layer as a primitive operation for end-to-end learning systems. The LML layer provides a probabilistic way of modeling multi-label predictions limited to having exactly $k$ labels. 

\textbf{Connect with Adversarial Attacks}. Since the loss with the top-$k$ operator is hard to be optimized, the authors in $k$Fool \cite{tursynbek2022geometry} and CW$^k$ \cite{zhang2020learning} use the sum of top-$k$ (a variant of AT$_k$ operator) to replace the top-$k$ operator. Hu et al., in \cite{hu2021tkml} propose a TKML-AP-T attack loss based on the sum of top-$k$, which aims to coerce the TKML classifier to use a specific set of $k$ labels that are not true classes of the input as the top-$k$ predictions. The idea of TKML-AP-T is also extended and further studied by Qaraei et al., in \cite{qaraei2021adversarial} for adversarial targeted attacks for extreme multilabel text classification with application in recommendation systems and automatic tagging of web-scale documents.
Hu et al. \cite{hu2021tkml} also replace the top-$k$ operator in the original TKML-AP-U and TKML-AP-Uv attacks with the average top-$k$ operator because they realized the average of top-$k$ is a tight upper-bound of the top-$k$ and the average of top-$k$ can be reformulated as a minimization problem, which is very easy to be optimized.  

\smallskip
\noindent
\shu{\textbf{Discussion}. The AT$_k$ individual loss can relax the Top-$k$ individual loss to be a convex loss. However, it may not account for the significance of each prediction and can be vulnerable to the wrong labels of the sample.}

\vspace{-1mm}
\begin{table}[h]
\centering
\scalebox{0.80}{
\begin{tabular}{l|l|l}
\hline
Work & Connection & Form \\ \hline
\cite{lapin2015top} & \makecell[l]{Multi-class}& \makecell[l]{$\frac{1}{k}\sum_{j=1}^k[1+f_{[j]}(\x)-f_{\y}(\x)]_+$} \\ \hline
\cite{fan2020groupwise}& \makecell[l]{Multi-label}& \makecell[l]{$\frac{1}{k}\sum_{j=1}^kf_{[j]}(\x)-\frac{1}{|\y|}\sum_{y\in\y}f_y(\x)$}\\ \hline
\cite{hu2021tkml}& \makecell[l]{Adversarial \\Attack}& \makecell[l]{$\|\delta\|_2^2+\sum_{j=1}^kf_{[k]}(\x+\delta)-\sum_{j\in\tilde{\y}}f_{j}(\x+\delta)$, where $\delta$ is a \\ perturbation. $\tilde{\y}\subset \mathbb{N}_C$ contains $k$ labels and $\tilde{\y}\cap\y=\oldemptyset$.}\\ \hline
\end{tabular}
}
\vspace{-0.5em}
\caption{\em \shu{Typical loss functions of the AT$_k$ individual loss.}
}
\label{table:ATk-individual}
\vspace{-0.5cm}
\end{table}

\vspace{-1mm}

\section{Connections with Other Topics}\label{sec:other_topics}
\vspace{-1mm}
In this section, we explore the relationship between rank-based losses and related topics in machine learning, such as data subset selection, hard example mining, DRO, and CVaR. These connections demonstrate the versatility and popularity of rank-based losses, and offer techniques for optimizing and designing algorithms that use them. 

\begin{itemize}[leftmargin=*]
    \item \textbf{Data subset selection} \cite{wei2015submodularity}. The data subset selection is a related topic in machine learning. It involves choosing a smaller subset from a larger training dataset to minimize average loss when training a model.  For example, curriculum learning \cite{bengio2009curriculum} and self-paced learning \cite{kumar2010self} are learning schemes that organize the training process into iterative passes, gradually including data from easy to hard-to-learn samples, measured by individual losses. As such, each pass of training in these methods corresponds to an average aggregate loss over the selected subset. Coreset\cite{borsos2020coresets} is another method of data subset selection. Solving the problem on the selected subset usually yields similar results as solving the same problem on the original dataset. It is an efficient method for handling massive datasets and streams, especially for online learning. These methods are related to the aggregate losses discussed earlier. For instance, selecting a subset of samples with the largest individual losses is related to average top-$k$ aggregate loss. 
    
    \item \textbf{Hard example mining}. Currently, training deep neural networks usually applies the hard example mining approach \cite{shrivastava2016training, lin2017focal}, which is similar to rank-based aggregate losses. For instance, \cite{shrivastava2016training} proposes online hard example mining that updates model parameters by using the top-$k$ samples (\ie, object proposal regions of interest) with the largest losses for each training image in object detection. Weighted cross-entropy loss, proposed in \cite{lin2017focal}, assigns more weight to candidate bounding boxes with large losses and demonstrated better suitability for object detection than classical cross-entropy loss.  However, it is important to note that the connection of these works to aggregate loss is incidental.
    
    \item \textbf{Distributionally robust optimization (DRO)}. DRO \cite{rahimian2019distributionally} is a popular method for robust learning that deals with data uncertainty from a probabilistic perspective. Its main aim is to ensure that a trained model performs well on unseen data. During training, it minimizes the worst-case weighted expected loss function on all training data, where the weight follows a probabilistic ambiguity set around the actual training data distribution. Aggregate loss strongly connects with DRO as discussed in Section \ref{sec:atk}, \eg, the average top-$k$ aggregate loss can be reformulated as a DRO problem. However, DRO suffers from instability during training, leading to poor performance on datasets containing outliers.  To address this issue,  \cite{zhai2021doro} proposes a robust outlier refinement of DRO called DORO. It adaptive removes a small fraction of examples with the largest individual losses during training, similar to the AB$_k$ and AoRR aggregate losses.
    
    \item \textbf{Condition Value at Risk (CVaR)}. Rank-based losses can be connected to CVaR \cite[Chapter~6]{ shapiro2021lectures}, which is a popular statistic used in risk-sensitive machine learning. It was originally introduced in economics research as a tool for quantifying the risk associated with a portfolio \cite{rockafellar2000optimization} and has many applications in operations research and machine. In machine learning, there are tasks that require models with good tail performance, i.e., models that perform well on worst-off examples in the dataset. CVaR loss computes the average risk over the tails of the loss and is well-suited for such scenarios.  This property is similar to the AT$_k$ aggregate loss, which also focuses on the largest individual losses. The work of \cite{fan2017learning} was the first to connect these two topics based on \cite{ogryczak2003minimizing}, and subsequently, \cite{hu2021sum} extended this connection to AoRR aggregate loss. 
    
\end{itemize}
\vspace{-3mm}



\section{Future Directions}\label{sec:future_directions}
\vspace{-1mm}
In this section, we suggest six directions that can be explored in the future.

\vspace{-3mm}

\begin{table*}[t]
\centering
\scalebox{0.90}{
\begin{tabular}{|c|c|c|}
\hline
Aggregators & Aggregate Losses & Individual Losses \\ \hline\hline
Average & \makecell[l]{
$\cdot$\textbf{Empirical Risk Minimization (ERM)} \cite{vapnik1999nature}, \cite{bartlett2006convexity, de2005model, steinwart2003optimal, wu2006learning}
\\ $\cdot$\textbf{Maximum (Log)-Likelihood} \cite{friedman2001elements}
}  
& \makecell[l]{$\cdot$\textbf{Multi-class Classification} \\ \ \ --\textit{All-in-One} \cite{weston1999support, bredensteiner1999multicategory, vapnik1999nature}; \textit{One-versus-All} \cite{vapnik1999nature}; \textit{LLW} \cite{lee2004multicategory}; \textit{RM-SVM}\cite{liu2011reinforced} \\ $\cdot$\textbf{Multi-Label Classification} \\ \ \ --\textit{Instance-F1 Loss} \cite{zhang2013review}; \textit{Hamming Loss}\cite{schapire2000boostexter, zhang2013review}}\\ \hline

Maximum & \makecell[l]{$\cdot$\textbf{Support Vector Machine (SVM)} \cite{clarkson2012sublinear, hazan2011beating}
\\$\cdot$\textbf{Robust Learning}  \cite{shalev2016minimizing, ben2002robust, namkoong2016stochastic}
\\$\cdot$\textbf{Minimax Learning} \cite{lan2020first, chen2017robust, carmon2021thinking}
\\$\cdot$\textbf{Federated Learning}  \cite{mohri2019agnostic}} & \makecell[l]{$\cdot$\textbf{Multi-class Classification} \\ \ \ --\textit{Multi-class SVM (Grammer and Singer)}\cite{crammer2001algorithmic}; \textit{AMO}\cite{dogan2016unified}; \textit{ATM}\cite{dogan2016unified} \\ $\cdot$\textbf{Multi-Label Classification} \\ \ \ --\textit{Conventional Multi-label Loss (Grammer and Singer)} \cite{crammer2003family}; \textit{MMP} \cite{furnkranz2008multilabel}; \\ \ \ \ \textit{Separation Ranking Loss} \cite{guo2011adaptive}; \textit{One-error Loss and Coverage Loss} \cite{schapire2000boostexter, zhang2013review}
 \\ $\cdot$\textbf{Adversarial Attacks} \\ \ \ --\textit{DeepFool} \cite{moosavi2016deepfool}; \textit{UAP} \cite{moosavi2017universal}; \textit{CW} \cite{carlini2017towards};  \textit{Multi-label Attacks} \cite{song2018multi, zhou2020generating, melacci2020can}
} \\ \hline

Median & \makecell[l]{$\cdot$\textbf{Least Median Squares (LMS)} \cite{shibzukhov2021principle, shibzukhov2022minimizing, rousseeuw1984least, rousseeuw1984robust, ma2011robust}
\\$\cdot$\textbf{Median-of-Means (MOM)} \cite{nemirovskij1983problem} \\ \ \ --\textit{Robust Regression} \cite{brownlees2015empirical, hsu2016loss, jalal2020robust, lecue2020robust, lugosi2019risk}; \textit{Imitation Learning} \cite{liu2022robust}}  & -- \\ \hline

Top-$k$ & -- & \makecell[l]{$\cdot$\textbf{Multi-class Classification} \\ \ \ --\textit{Top-$k$ SVM} \cite{lapin2015top, yan2018top, zhang2019deep, oh2017top, chzhen2021set, tan2019exact}; \textit{Top-$k$ Calibrated Loss} \cite{yang2020consistency}; \\ \ \ \  \textit{Smoothed Top-$k$ Calibrated Loss} \cite{garcin2022stochastic}
 \\ $\cdot$\textbf{Multi-Label Classification} \cite{li2017improving} --\textit{TKML} \cite{hu2020learning}; \textit{TKML-AoRR}\cite{hu2021sum}
 \\ $\cdot$\textbf{Adversarial Attacks} \\ \ \ --\textit{$k$Fool}\cite{tursynbek2022geometry}; \textit{CW$^k$}\cite{zhang2020learning}; \textit{$k$UAPs} \cite{tursynbek2022geometry}; \textit{TKML-AP-U}\cite{hu2021tkml}; \textit{TKML-AP-Uv}\cite{hu2021tkml}
} \\ \hline

AT$_k$ & \makecell[l]{$\cdot$\textbf{Conditional Value at Risk (CVaR)} \cite{fan2017learning}, \cite{williamson2019fairness, curi2020adaptive, lee2020learning, laguel2021superquantiles, holland2021spectral, laguel2022superquantile} \\ \ \ --\textit{Fairness} \cite{williamson2019fairness}; \textit{Adversarial Training} \cite{robey2022probabilistically} \\$\cdot$\textbf{SVM} --\textit{C-SVM} \cite{cortes1995support}; \textit{$\nu$-SVM} \cite{scholkopf2000new}; \textit{E$\nu$-SVM}\cite{perez2003extension, takeda2008nu} \\ $\cdot$\textbf{Distributionally Robust Optimization (DRO)}\cite{lyu2020average, levy2020large, roux2021efficient} \\ \ \ --\textit{Active Learning}\cite{zhu2019robust}; \textit{DRO-TopK}\cite{qi2020simple}; \\ \ \ \ \textit{Multiple Instance Learning}\cite{sapkota2021distributionally} \\ $\cdot$\textbf{Learning Strategies} \\ \ \ --\textit{Ordered SGD}\cite{kawaguchi2020ordered}; \textit{Large Current Loss Minimization}\cite{wu2020curricula}; \\ \ \ \ \textit{Tilted ERM}\cite{li2020tilted, li2021tilted}; \textit{AT$_k$-GSAM}\cite{yuan2020group} \\ $\cdot$\textbf{Deep Learning Applications} \\ \ \ --\textit{Optic Disc and Optic Cup Segmentation}\cite{xu2019mixed}; \\ \ \ \ \textit{Sonar Image Generation}\cite{lee2018deep,lee2019deep};\\ \ \ \  \textit{Head Pose Estimation}\cite{huang2020improving}; \textit{6D Pose Estimation}\cite{jeon2020prima6d}; \\ \ \ \ \textit{ChIP-seq Identification}\cite{oh2020cnn}; \textit{Multi-task  Learning}\cite{wang2021simultaneous};  \\ \ \ \ \textit{Top-k Generative Adversarial Network (GAN)}\cite{sinha2020top}}
 & \makecell[l]{$\cdot$\textbf{Multi-class Classification} \\ \ \ --\textit{(Smoothed) Top-$k$ SVM$^{\alpha}$}\cite{lapin2015top, chu2018optimizing, lapin2016loss, lapin2017analysis, berrada2018smooth, zhang2019deep}; \\ \ \ \  \textit{(Smoothed) Top-$k$ SVM$^{\beta}$}\cite{lapin2015top, lapin2016loss, lapin2017analysis, rawat2020doubly}; \textit{Robust Top-$k$ SVM} \cite{chang2017robust}; \\ \ \ \ \textit{Convex Top-$k$ Partial Loss} \cite{gong2021top}; \textit{(LDR)-$k$} \cite{zhu2021unified};\\ \ \ \  \textit{Weighted Top-$k$ SVM$^{\alpha}$ and Weighted Top-$k$ SVM$^{\beta}$ }\cite{kato2019learning, tajima2021frank}; \\ \ \ \   \textit{OWLs}\cite{reddi2019stochastic};  \textit{(Truncated) Top-$k$ Entropy Loss} \cite{lapin2016loss, lapin2017analysis, zhao2019appusage2vec, furnari2018leveraging}; \\ \ \ \ \textit{Top-$k$ Grouping Loss}\cite{sawada2020trade}
 \\ $\cdot$\textbf{Multi-Label Classification}\cite{amos2019limited} \\ \ \ --\textit{GRLS} \cite{fan2020groupwise}
 \\ $\cdot$\textbf{Adversarial Attacks} \cite{qaraei2021adversarial} \\ \ \ --\textit{$k$Fool} \cite{tursynbek2022geometry}; \textit{CW$^k$}\cite{zhang2020learning}; \\ \ \ \ \textit{TKML-AP-T} \cite{hu2021tkml}; \textit{TKML-AP-U} \cite{hu2021tkml}; \textit{TKML-AP-Uv} \cite{hu2021tkml}
}  \\ \hline
 
AB$_k$ & \makecell[l]{$\cdot$\textbf{Trimmed Loss} \cite{yuan2020learning, song2020no, gui2021towards}\\ \ \ --\textit{Least Trimmed Square (LTS)} \cite{rousseeuw1984least}; \textit{Iterative LTS} \cite{shen2019learning}; \\ \ \ \ \textit{GAN} \cite{shen2019learning}; \textit{Mixed Linear Regression} \cite{shen2019iterative}; \textit{MKL-SGD} \cite{shah2020choosing} \\ $\cdot$\textbf{Fairness} \cite{roh2021sample}
  $\cdot$\textbf{Inverted CVaR} \cite{lee2020learning, han2018co} \\ $\cdot$\textbf{Robust Unsupervised 
Learning} \cite{maurer2021robust}} & -- \\ \hline

AoRR & \makecell[l]{$\cdot$\textbf{Bilevel Optimization} \cite{hu2021sum}  $\cdot$\textbf{Interval CVaR} \cite{hu2021sum, liu2022risk} \\$\cdot$\textbf{Difference of Convex Optimization} \cite{hu2020learning, yao2022large} \\$\cdot$\textbf{SVM} \\ \ \ --\textit{ER-SVM} \cite{fujiwara2017dc}; \textit{Ramp-loss 
SVM} \cite{collobert2006trading}; \\ \ \ \ \textit{Robust Outlier Detection (ROD)} \cite{xu2006robust}; \\ \ \ \ \textit{CVaR-($\alpha_L$, $\alpha_U$)-SVM} \cite{tsyurmasto2013support}; \textit{($\nu$, $\mu$)-SVM} \cite{kanamori2017breakdown, kanamori2017robustness}  \\$\cdot$\textbf{Trimmed Root Mean Square Error (tRMSE)}\cite{ortis2019predicting} } & -- \\ \hline
 
Close-$k$ & \makecell[l]{$\cdot$\textbf{Robust Learning} \cite{he2018minimizing}}  & -- \\ \hline
\end{tabular}
}
\caption{\em An overview of related references on rank-based decomposable aggregate losses and rank-based decomposable individual losses. The empty block means the current aggregator in the corresponding aggregate losses/individual losses field has not been explored. The major topics are shown in bold. The sub-topics (in the Aggregate Losses column) and the existing losses (in the Individual Losses column) are shown in italic.
}
\label{table:summary}
\vspace{-0.5cm}
\end{table*}

\subsection{New Aggregators and Nested Aggregators}
\vspace{-1mm}
Table \ref{table:summary} summarizes the previous works on different types of aggregate loss and individual loss with various aggregators.  We observe that none of the previous works have utilized the top-$k$ aggregator for designing aggregate losses, and four aggregators (Median, AB$_k$, AoRR, and Close-$k$) have not been explored in individual losses. \shu{Each aggregator has its own strengths and may be suitable for specific tasks and scenarios when designing loss functions. Therefore, further research on these missing aggregators in aggregate or individual losses can be valuable. However, to identify and justify the usefulness of new aggregators for designing loss functions, we need to study the problems they can solve and analyze how different aggregators can effectively address them. This can be challenging.}

Fig. \ref{fig:aggregators} shows the relationships between all aggregators, and it suggests that the AoRR aggregator is the most general among them. However, this also indicates that there is potential to develop new aggregators that are even more general than AoRR. One promising approach for designing such aggregators is to study the ranking relationships between individual values. Therefore, further exploration of new aggregators is meaningful and may lead to improved performance in specific scenarios and tasks.

Furthermore, we can combine rank-based aggregate and individual losses in various ways.  Two existing works \cite{rawat2020doubly, hu2021sum} have explored this combination. Specifically, \cite{rawat2020doubly} uses AT$_k$ aggregator on both aggregate and individual loss levels, while \cite{hu2021sum} applies AoRR aggregator for aggregate loss level and Top-$k$ aggregator for individual loss level. Therefore, at least $7\times 4-2 = 26$ possible combination methods can be tried, excluding these two existing methods, where 7 and 4 are the number of aggregators that have existed in aggregate loss and individual loss, respectively. In addition, the combination can be applied only on one side.  For example, in meta-learning\cite{finn2017model}, the goal of a learning task is to learn a new task efficiently by meta-training a meta-model (called meta-learner) from a set of historical tasks (called base-learner when working on each historical task). In this case, the aggregator can be used to learn both the meta-learner and the base-learner.
These combined aggregators are called nested aggregators, which is an interesting topic for further exploration.

\vspace{-3mm}

\subsection{Geometric Interpretation of Rank-based Losses}
\vspace{-1mm}
\begin{figure}[t]
\vspace{-\intextsep}
\centering
\includegraphics[trim=1 1 1 1, clip,keepaspectratio, width=0.45\textwidth]{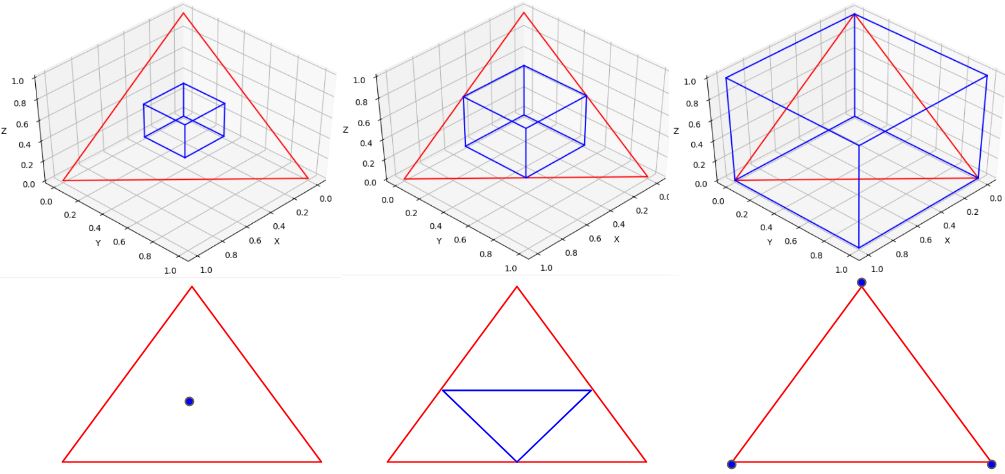}
\vspace{-0.2cm}
\caption{\it 3D (top) and 2D (bottom) interpretation of three aggregators. \textbf{Left}: average aggregator. \textbf{Middle}: AT$_k$ aggregator.  \textbf{Right}: maximum aggregator.}
\vspace{-0.5cm}
\label{fig:interpretation}
\end{figure}
To better understand rank-based losses, viewing them from a geometric perspective is helpful, but this approach has not been explored much in the current literature. 
In \cite{blondel2020learning}, Blondel et al. treat ranking as a structured prediction problem. Lapin et al. \cite{lapin2015top} regard the AT$_k$ simplex as a convex polytope in multi-class SVM for individual loss. Similarly, Kong et al. \cite{kong2020rankmax} interpret the softmax function as a projection on the ($n$, $k$)-simplex. Furthermore, Lyu et al. \cite{lyu2020average} discuss the interpretation of AT$_k$ aggregate loss based on its connection with DRO. All of them consider the aggregators (\ie, average, maximum, and AT$_k$) as constrained learning problems:
\vspace{-2mm}
\begin{equation*}
    \begin{aligned}
    &\mathcal{L}(\ell(\mathcal{D})):= \max_{\w}\sum_{i=1}^n \w_i\ell_i \ \text{or} \  \mathcal{L}(\mathbb{O}(f(\x)))=\max_{\w}\sum_{i=1}^n \w_i \mathcal{O}_i,\\
    &\mbox{s.t.} \ \w^\top \mathbf{1}= 1, \w\geq 0, \ \|\w\|_{\infty}\leq \epsilon, \mbox{with} \ \epsilon\in \left[1/n, 1\right]. 
    \end{aligned}    
\end{equation*}
\vspace{-1mm}
where $\mathcal{O}_i =\mathcal{O}(f_{i}(\x),\y_i)$.

From the above constraints, we know $\w$ belongs to a n-dimensional probability simplex $\Delta_n$ such that $\Delta_n:=\{\w\in \mathbb{R}_+^n: \w^\top \mathbf{1}= 1\}$. On the other hand, $\w$ is within an $l_\infty$-ball centered at $\frac{1}{n}\mathbf{1}$ with radius $\epsilon>0$. Here are some examples when $n=3$:
\begin{itemize}[leftmargin=*]
    \item If $\epsilon=1/3$, the learning objective will generalize to the average loss, which is shown in Fig. \ref{fig:interpretation} (left).
    \item If $\epsilon=1/2$, the learning objective becomes the average top 2 losses. Fig. \ref{fig:interpretation} (middle) shows its geometry interpretation. Note that such constraints are sometimes called the capped probability simplex \cite{lim2016efficient} as mentioned in \cite{blondel2020learning}.
    \item If $\epsilon=1$, the learning objective will be reduced to the maximum loss. See Fig. \ref{fig:interpretation} (right) for more details.
\end{itemize}
Using a geometric interpretation is a helpful way to understand rank-based losses and aggregators, but it's not yet clear how to apply this approach to other aggregators like AB$_k$, AoRR, and close-$k$. This is still an open question.
\vspace{-3mm}

\subsection{Converting Non-decomposable Aggregate Losses to Decomposable Aggregate Losses}
\vspace{-1mm}

Certain sensitive domains, such as medicine and biometrics, require using \textit{non-decomposable} loss functions \cite{ranjbar2012optimizing, eban2017scalable}. In particular, in the field of Information Retrieval (IR), they are commonly used \cite{liu2009learning, li2011short}, 
These losses have the advantage of being able to handle data with label imbalance. However, optimizing models using these losses to perform well on specific evaluation metrics can be challenging because they are often not decomposable across samples. This can lead to computational scalability issues. To address this problem, some researchers convert non-decomposable rank-based aggregate losses to decomposable ones, which can achieve the same learning goals but with more efficiency and effectiveness in optimization procedures. 

For example, Eban et al. \cite{eban2017scalable} give convex relaxations for information retrieval metrics with decomposable objectives, leading to training that scales to large datasets. Boyd et al. in \cite{boyd2012accuracy} propose a method for measuring classification accuracy at the top $\tau$-quantile values of a scoring function such that $\tau\in[0,1]$. They propose AATP loss to push the examples whose scores are among the top $\tau$-quantile are as relevant as possible in an ordering of all examples, which can be defined as, $\mathcal{L} (f, \mathcal{D}) := \frac{1}{|\mathcal{D}|}\left[\sum_{i\in \mathcal{D}^+}\mathbb{I}_{f(x_i^+)< q_f}+\sum_{j\in \mathcal{D}^-}\mathbb{I}_{f(x_i^-)> q_f}\right],$
where $q_f$ is the top $\tau$-quantile of the values taken by $f$, $|\mathcal{D}|$ is the total number of examples, and $\mathcal{D}^+$ and $\mathcal{D}^-$ represent the positive and negative example index set, respectively. 
Lyu et al. \cite{lyu2018univariate} introduce a decomposable surrogate loss for AUC loss, which subtracts the sum of prediction scores of all positive examples from the sum of the top-$|\mathcal{P}|$ prediction scores in the prediction score list, where $|\mathcal{P}|$ is the total number of positive examples. 

All of these methods can only handle some simple metrics, but many complicated metrics have not been well studied and explored in connection with decomposable aggregate losses. For example, the average precision (AP) \cite{baeza1999modern}, discounted cumulative gain (DCG) \cite{jarvelin2017ir, jarvelin2002cumulated}, partial AUC \cite{narasimhan2013svmpauctight}, pAp@$k$ \cite{hiranandani2020optimization}, Mean reciprocal rank (MRR) \cite{craswell2009mean}, ERR-IA \cite{chapelle2009expected}, rank aggregation \cite{dwork2001rank}, etc. 
\vspace{-3mm}

\subsection{Statistical Machine Learning Theory on Rank-based Losses}
\vspace{-1mm}
To design a loss function, we need to analyze its statistical properties, and then develop an efficient algorithm to optimize it. There are three key factors that must be considered when justifying the applicability of a designed loss function: classification calibration and consistency, generalization capability, and optimization error or convergence guarantee. We will now discuss each of these factors in detail.
\begin{itemize}[leftmargin=*]
    \item \textbf{Classification Calibration and Consistency}. To design a robust model in classification learning theory, we need to start by designing a surrogate loss because the 0/1 loss, which represents the optimal Bayes error, is an NP-hard problem for most hypothesis sets. The consistency property of the surrogate loss is essential, which means that the optimal minimizers of the surrogate loss should be near or equal to the optimal minimizers of the 0/1 loss \cite{awasthi2021calibration}. Consistency has been widely studied in binary \cite{zhang2004statistical, bartlett2006convexity}, multi-class \cite{tewari2007consistency}, and multi-label classification \cite{gao2011consistency}. Another related concept is classification calibration, which is a sufficient condition for consistency. A surrogate loss is considered calibrated if the model's predicted probabilities match the empirical frequencies, which helps to avoid incorrect decisions when the outcome is uncertain \cite[Chapter~13.2.2]{ murphy2023probabilistic}.
    
    Several studies have investigated the consistency and calibration properties of the average aggregate loss  \cite{zhang2004statistical, bartlett2006convexity}, AT$_k$ aggregate loss \cite{fan2017learning, lyu2020average}, and AoRR aggregate loss \cite{hu2020learning, hu2021sum}.  However, the consistency and calibration properties of Median, AB$_k$, and Close-$k$ aggregate losses are still unclear, except for the calibration property of the Maximum aggregate loss, which has been discussed in \cite{he2018minimizing}. 
    On the other hand, for Average and Maximum guided individual losses, \cite{tewari2007consistency, gao2011consistency} have explored their consistency and calibration properties based on the binary case results, but the properties of Top-$k$ and AT$_k$ guided individual losses are challenging to analyze. To overcome this, \cite{lapin2016loss, lapin2017analysis} propose the notion of top-$k$ calibration, and \cite{yang2020consistency} defines a new top-$k$ consistency based on top-$k$ calibration under certain conditions. However, these definitions are only for multi-class classification, and no existing works discuss top-$k$ calibration and consistency for multi-label learning, such as for TKML loss \cite{hu2020learning}. This is a potential direction for future research. 
    
    \item \textbf{Generalization Capability}. In machine learning, we use a finite training dataset to find optimal values for the model parameters, and then apply this model to new test datasets that are assumed to be identically distributed. However, in real-world situations, two finite datasets sampled from the same distribution can have different characteristics, leading to poor generalization. The difference between the errors on the training and test datasets is known as the generalization gap. The objective of learning is to minimize this gap and improve the model's generalization capability.

    The average aggregate loss is known as empirical risk minimization (ERM). It has been extensively studied for its generalization capability through Vapnik–Chervonenkis (VC) dimension in Probably Approximately Correct (PAC) learning \cite{shalev2014understanding} or Rademacher complexity \cite{mohri2018foundations}. The generalization capability of maximum aggregate loss is also studied  in \cite{shalev2016minimizing} based on the VC dimension. \cite{fan2017learning, lyu2020average} connect AT$_k$ to SVM and then analyze its generalization bound. In addition, AB$_k$'s generalization bound is studied in 
    \cite{lee2020learning}. Hu et al. \cite{hu2021sum} provide a generalization bound for AoRR aggregate loss based on CVaR. 
    However, the generalization capability of Median and Close-$k$ aggregate losses is not well-studied.  
    
    \item \textbf{Optimization Error (Convergence Guarantee) of Learning Strategies}. To optimize a loss function, we need an efficient algorithm, which can greatly affect the final model's performance. In addition, loss functions are not always convex, making it difficult for algorithms to find the global optima. Therefore, it's important to study the optimization error and convergence guarantee of learning strategies combined with the loss function.
    
    Aggregate losses such as Average, Maximum, Median, AT$_k$, AB$_k$, and AoRR are usually optimized using gradient or sub-gradient-based descent or ascent algorithms. 
    Optimization errors of these algorithms are studied under specific conditions and tasks. For instance, optimization methods \cite{curi2020adaptive, namkoong2016stochastic} used for CVaR and their optimization errors can be adapted for AT$_k$ aggregate loss, while the optimization error on AoRR \cite{hu2021sum} is directly dominated by the difference-of-convex algorithm (DCA) \cite{tao1986algorithms}. 
    However, the optimization error for Close-$k$ aggregate loss has not been explored in the current literature. For most of the rank-based individual losses, they are optimized by using popular algorithms with existing convergence guarantee (\eg, average guided \cite{lee2004multicategory, liu2011reinforced}, maximum guided \cite{guo2011adaptive}, top-$k$ guided \cite{lapin2015top}, and average top-$k$ guided \cite{lapin2015top}.) or heuristic algorithms without convergence guarantee (\eg, average guided \cite{bredensteiner1999multicategory}, maximum guided \cite{crammer2001algorithmic, crammer2003family}, top-$k$ guided \cite{yan2018top}, and average top-$k$ guided \cite{fan2020groupwise}.). Therefore, exploring effective algorithms with low optimization errors for solving rank-based aggregate losses and individual losses is significant and worthwhile in the future. 
\end{itemize}
\vspace{-3mm}

\subsection{Aggregate Gradients} 
\vspace{-1mm}
Aggregate losses are often associated with gradients since optimizing the selected individual losses is equivalent to selecting their corresponding gradients to update the model parameters during training.  Importance sampling \cite{katharopoulos2017biased, katharopoulos2018not, johnson2018training} techniques suggest that examples should be sampled with probability proportional to the norm of the loss term's gradient. Therefore, the gradient norm can be a useful cue to select necessary samples, speed up SGD-based algorithms, and reduce the stochastic gradient's variance.  Paul et al. in \cite{paul2021deep} find that the loss gradient norm of individual training samples can be used to identify a smaller subset of the training dataset that can benefit the model's generalization during the beginning of the training.  Moreover, the gradient norm can be used to prune a significant fraction of the training set without sacrificing the model's generalization ability after a few training epochs. In addition, the gradient norm is further studied in \cite{liu2021learning} for training DNNs in the presence of corrupted supervision. The proposed method in \cite{liu2021learning} focuses on controlling the collective impact of data points on the average gradient. They observed that the gradient norm-based method is better than the loss value-based for handling the corrupted data in experiments.  Therefore, future research can explore learning with a rank-based gradient norm, and studying aggregators on gradient norms can be an exciting topic.
\vspace{-3mm}

\subsection{Hyperparameter Tuning}
\vspace{-1mm}
Rank-based losses rely heavily on important hyperparameters that can greatly affect the performance of the final model. For instance, in the case of AoRR aggregate loss, the hyperparameter $m$ determines the number of possible outliers that are ignored during training. If $m$ is set lower than the actual number of outliers, the model will be adversely affected by the outliers. Conversely, if $m$ is set too high, some essential samples will be removed during training, resulting in a suboptimal final model. 

Several works have attempted to develop new learning strategies to determine the hyperparameters in model training. For example, Kawaguchi et al. used the AT$_k$ aggregator to design ordered SGD optimization methods in \cite{kawaguchi2020ordered}. They employed an adaptive setting to decrease $k$ when the model performance achieved preset-specific criteria. In \cite{hu2020learning}, a similar method was applied for learning the AoRR aggregate loss. They used greedy search to determine the hyperparameter $m$. However, these methods may not be efficient for large-scale datasets. Therefore, in \cite{hu2021sum}, Hu et al. proposed an auxiliary learning framework to determine the hyperparameters using a clean dataset, which is extracted from the original dataset that may contain outliers. However, this method may not work when a clean dataset is unavailable. Therefore, exploring methods for determining hyperparameters in rank-based decomposable losses may be a promising future direction.     
\vspace{-3mm}





\section{Conclusion}\label{sec:conclusion}
\vspace{-1mm}
This paper provides the first systematic and comprehensive survey of rank-based decomposable losses in machine learning. Concretely, we define a rank-based set function named aggregator and apply it to formulate aggregate and individual losses. We review existing literature on rank-based decomposable aggregate and individual losses, and show how they are connected to various research topics in machine learning. We further suggest six potential directions for future research. We find that the rank-based decomposable losses are indeed interesting to many researchers from different fields in machine learning.  We believe this survey can push forward future research in this area and hope to help researchers to apply these existing losses and investigate new rank-based decomposable losses.  
\vspace{-3mm}



\ifCLASSOPTIONcaptionsoff
  \newpage
\fi



%



\bibliographystyle{IEEEtran}
\bibliography{main_PAMI}
\vspace{-1.0cm}

\begin{IEEEbiography}
[{\includegraphics[width=1.1in,height=1.25in]{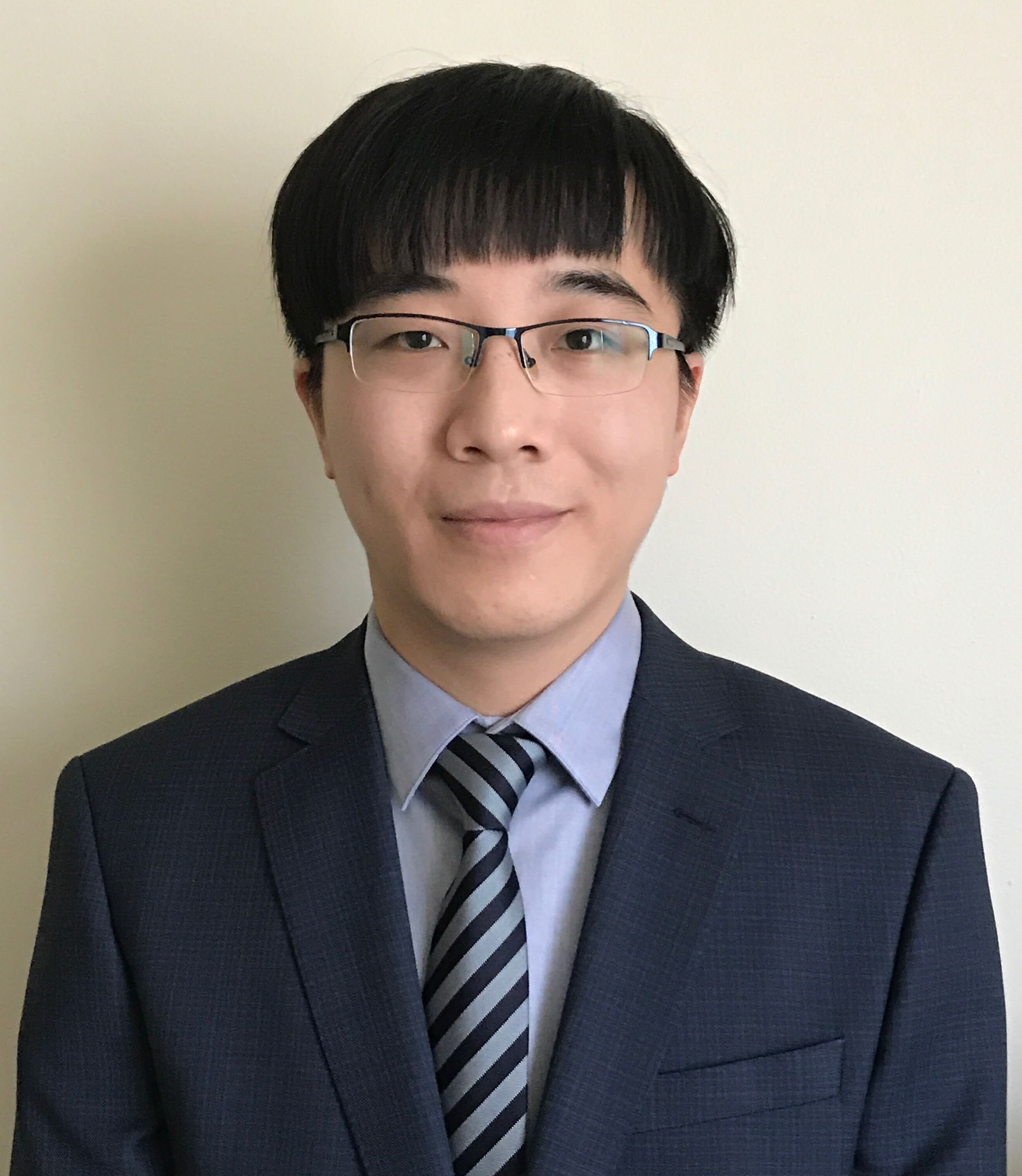}}]
{Dr. Shu Hu} is an Assistant professor in the Department of Computer Information and Graphics Technology within the Purdue School of Engineering and Technology at Indiana University-Purdue University Indianapolis. He was a Postdoc at Carnegie Mellon University. He received his Ph.D. degree in Computer Science and Engineering from University at Buffalo, the State University of New York (SUNY) in 2022. He received his M.A. degree in Mathematics from University at Albany, SUNY in 2020, and M.Eng. degree in Software Engineering from University of Science and Technology of China in 2016. His research interests include machine learning, digital media forensics, and computer vision.  
\vspace{-1.0cm}
\end{IEEEbiography}

\begin{IEEEbiography}
[{\includegraphics[width=1.1in,height=1.25in]{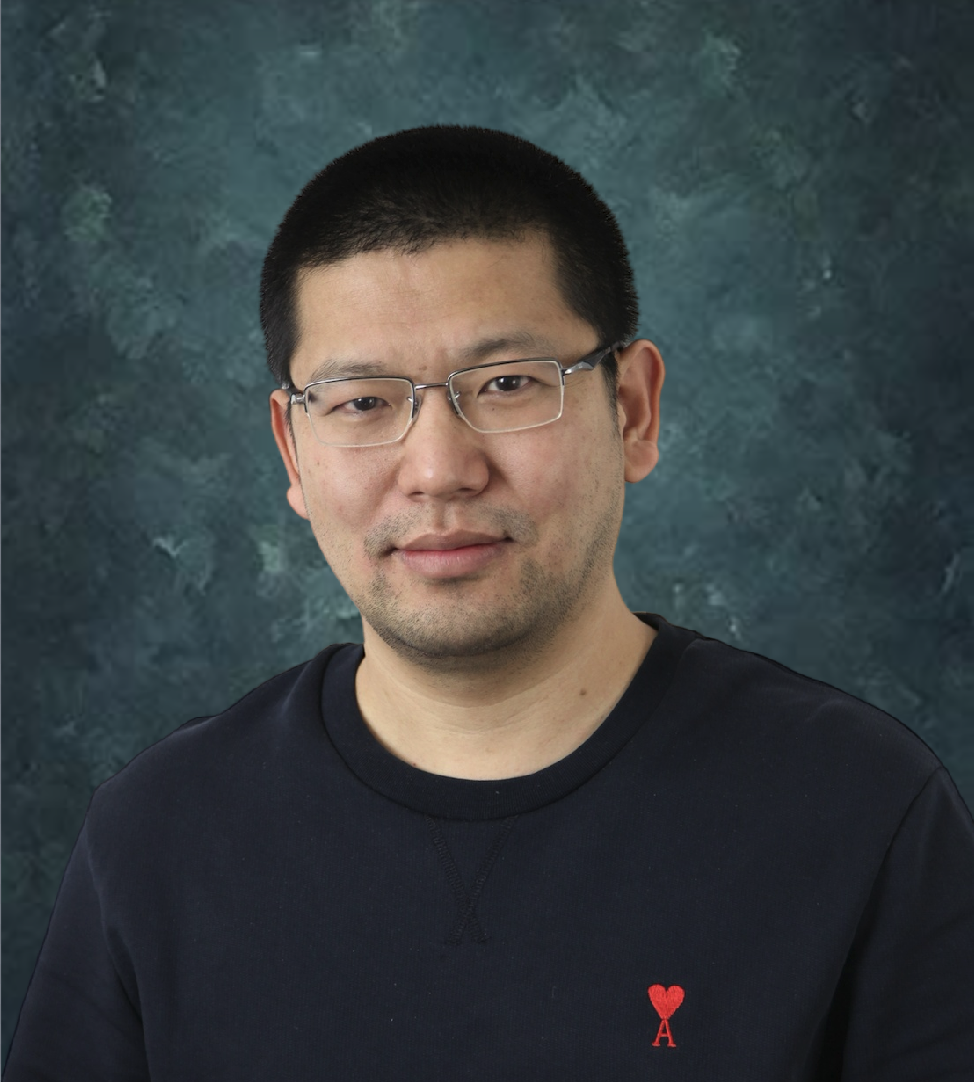}}]
{Dr. Xin Wang} (SM'2020)
is a Research Affiliate at University at Buffalo, State University of New York. He received his Ph.D. degree in Computer Science from the University at Albany, State University of New York in 2015. His research interests are in machine learning, reinforcement learning, deep learning, and their applications. He is a senior member of IEEE. 
\vspace{-1.0cm}
\end{IEEEbiography}

\begin{IEEEbiography}
[{\includegraphics[width=1.1in,height=1.25in]{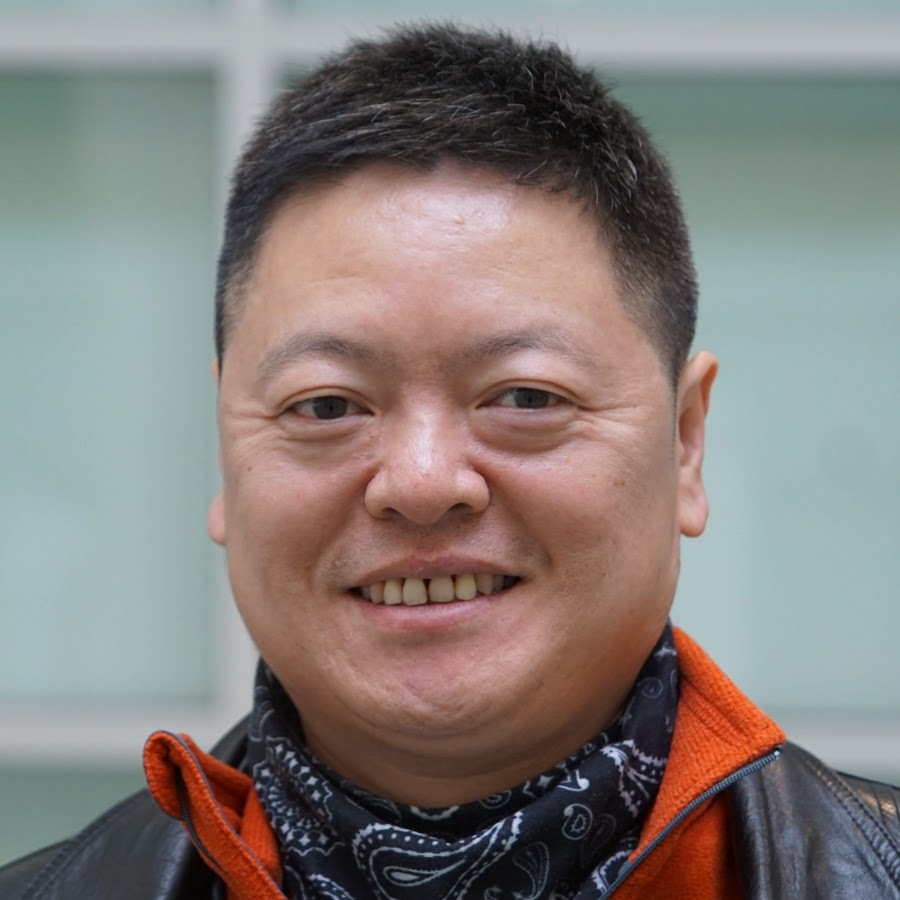}}]
{Dr. Siwei Lyu} is a SUNY Empire Innovation Professor at the Department of Computer Science and Engineering, of the University at Buffalo, State University of New York, USA. Dr. Lyu earned his Ph.D. in Computer Science from Dartmouth College in 2005 and both his M.S. (2000) and B.S. (1997) degrees in Computer Science and Information Science, respectively, from Peking University, China. Dr. Lyu’s research interests include digital media forensics, computer vision, and machine learning. He is a Fellow of IEEE, IAPR, and AAIA, and a Distinguished Member of ACM.
\end{IEEEbiography}

%








\end{document}